\documentclass[10pt,letterpaper,twocolumn]{article}

\usepackage[T1]{fontenc}
\usepackage[utf8]{inputenc}
\usepackage[margin=0.75in,top=0.75in,bottom=0.85in,columnsep=0.30in]{geometry}
\usepackage{newtxtext}
\usepackage{amsmath,amssymb,mathtools,bm}
\usepackage[table]{xcolor}
\usepackage{graphicx}
\usepackage[hyphens]{url}
\usepackage[round,authoryear]{natbib}
\let\cite\citep
\usepackage{multibib}
\newcites{app}{Appendix References}
\usepackage{caption}
\usepackage{booktabs,longtable,tabularx,array,multirow}
\usepackage{algorithm}
\usepackage{algpseudocode}
\usepackage{listings}
\usepackage{tcolorbox}
\tcbuselibrary{breakable,listings,skins}
\usepackage{microtype}
\usepackage[hidelinks,bookmarksnumbered]{hyperref}
\usepackage{bookmark}

\allowdisplaybreaks
\definecolor{bestgreen}{RGB}{46,125,50}
\definecolor{secondgreen}{RGB}{129,199,132}
\definecolor{thirdgreen}{RGB}{200,230,201}
\newcommand{\best}[1]{\cellcolor{bestgreen!85}\textcolor{white}{#1}}
\newcommand{\second}[1]{\cellcolor{secondgreen!70}#1}
\newcommand{\third}[1]{\cellcolor{thirdgreen!80}#1}

\newcolumntype{L}[1]{>{\raggedright\arraybackslash}p{#1}}
\newcolumntype{Y}{>{\raggedright\arraybackslash}p{0.72\linewidth}}
\lstdefinestyle{promptstyle}{
  basicstyle=\ttfamily\footnotesize,
  breaklines=true,
  columns=fullflexible,
  frame=single,
  showstringspaces=false,
  keepspaces=true,
  xleftmargin=1em,
  xrightmargin=1em
}

\title{Hallucination-aware Faithful Reasoning in Remote Sensing}
\definecolor{authoremailblue}{RGB}{64,137,195}
\newsavebox{\paperauthorbox}
\newcommand{\paperauthors}{%
  Yi Liu$^{1,2}$, Jing Zhang$^{1,2\dagger}$, Xiang Feng$^{1}$,
  Xiaoyu Tian$^{3}$, Haonan Guo$^{1,2}$, Di Wang$^{1,2}$,
  Zengmao Wang$^{1}$, Bo Du$^{1,2\dagger}$}
\newcommand{\paperaffiliations}{%
  $^{1}$School of Computer Science, Wuhan University,\quad
  $^{2}$Zhongguancun Academy,\quad \\
  $^{3}$School of Computer Science, Chongqing University}
\newcommand{\paperemails}{%
  \href{mailto:jingzhang.cv@gmail.com}{\textcolor{authoremailblue}{jingzhang.cv@gmail.com}}%
  \textcolor{red}{;}\quad
  \href{mailto:dubo@whu.edu.cn}{\textcolor{authoremailblue}{dubo@whu.edu.cn}}}
\author{\paperauthors}
\date{}
\hypersetup{
  pdftitle={Hallucination-aware Faithful Reasoning in Remote Sensing},
  pdfauthor={Yi Liu, Jing Zhang, Xiang Feng, Xiaoyu Tian, Haonan Guo, Di Wang, Zengmao Wang, Bo Du},
  pdfsubject={FaithRS: faithful reasoning in remote sensing},
  pdfkeywords={FaithRS, remote sensing, visual question answering, hallucination}
}

\makeatletter
\renewcommand{\@maketitle}{%
  \newpage\null
  \begin{center}
    {\LARGE\bfseries\@title\par}
    \vspace{0.8em}
    \ifx\paperauthors\@empty\else
      \sbox{\paperauthorbox}{{\large\paperauthors}}%
      \ifdim\wd\paperauthorbox>0.98\textwidth
        \resizebox{0.98\textwidth}{!}{\usebox{\paperauthorbox}}%
      \else
        \usebox{\paperauthorbox}%
      \fi\par
      \vspace{0.4em}
    \fi
    \ifx\paperaffiliations\@empty\else
      {\normalsize\begin{tabular}[t]{@{}c@{}}\paperaffiliations\end{tabular}\par}
      \vspace{0.2em}
    \fi
    \ifx\paperemails\@empty\else
      {\normalsize\ttfamily\paperemails\par}
    \fi
  \end{center}
  \begin{center}
    \captionsetup{hypcap=false}
    \includegraphics[width=0.98\textwidth]{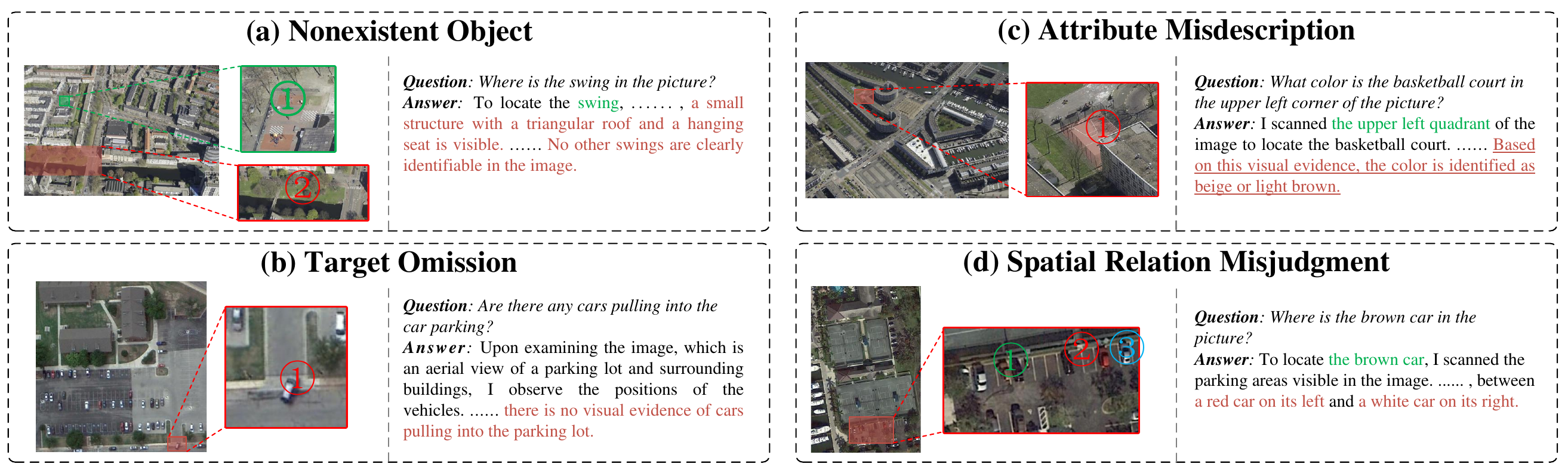}
    \captionof{figure}{Examples of typical process-level hallucinations in remote sensing visual question answering. (a) Nonexistent object: the model fabricates a swing-like structure with a triangular roof and a suspended seat in an area where no such object is visible. (b) Target omission: the model fails to recognize vehicles that are entering the parking lot. (c) Attribute misdescription: the model describes the surface color of the basketball court in a manner that is inconsistent with the image evidence. (d) Spatial relation misjudgment: the model incorrectly infers the relative position of the target vehicle.}
    \label{fig:motivation}
  \end{center}
  \vspace{0.5em}
}
\makeatother

\begin{document}
\maketitle
\begingroup
  \renewcommand{\thefootnote}{\ensuremath{\dagger}}
  \footnotetext[2]{Corresponding authors.}
\endgroup

% 0_abstract
\begin{abstract}
{Multimodal large language models (MLLMs) have demonstrated promising capabilities for remote sensing reasoning, yet they frequently generate visually unsupported reasoning that contradicts the visual evidence. Existing post-training methods optimize primarily for final-answer correctness, providing limited supervision for identifying and correcting hallucinations that arise during the reasoning process. To address this limitation, we propose \textbf{FaithRS}, a hallucination-aware self-reflection framework for faithful reasoning in remote sensing. We first construct \textbf{SFHD-43K}, a dataset of 43K structured, fine-grained hallucination annotations, and train a detector to identify hallucination types, underlying causes, and supporting visual evidence. FaithRS then incorporates this diagnostic feedback into a Safe Detector-Guided Corrective Rollout framework, where a safety-aware gate selectively determines whether and how to revise the response. Training is driven by a composite reward that jointly optimizes answer accuracy, reasoning faithfulness, hallucination reduction, format compliance, and revision safety. Experiments with Qwen3.5 backbones (0.8B, 4B, and 9B) show consistent improvements across four remote sensing benchmarks, with gains of up to 11.64 points. Code and data will be publicly available at \href{https://github.com/MiliLab/FaithRS}{\texttt{https://github.com/MiliLab/FaithRS}}.}
\end{abstract}

% 1_intro
\section{Introduction}

{MLLMs~\cite{dai2023instructblip, liu2023visual, chen2024internvl, yang2025qwen3} have expanded the capability of remote sensing analysis beyond conventional recognition and classification toward open-ended image interpretation, visual question answering, object localization, and spatial reasoning~\cite{hu2023rsgpt, kuckreja2023geochat, luo2024skysensegpt}. However, large scale variation, ambiguous object boundaries, and dense spatial relations make remote sensing MLLMs particularly prone to visual hallucinations. As shown in Fig.~\ref{fig:motivation}(a), the model fabricates a swing-like structure in a region where no such object exists, undermining the reliability of its reasoning.}

\begin{figure*}[t]
    \centering
    \includegraphics[width=0.98\linewidth]{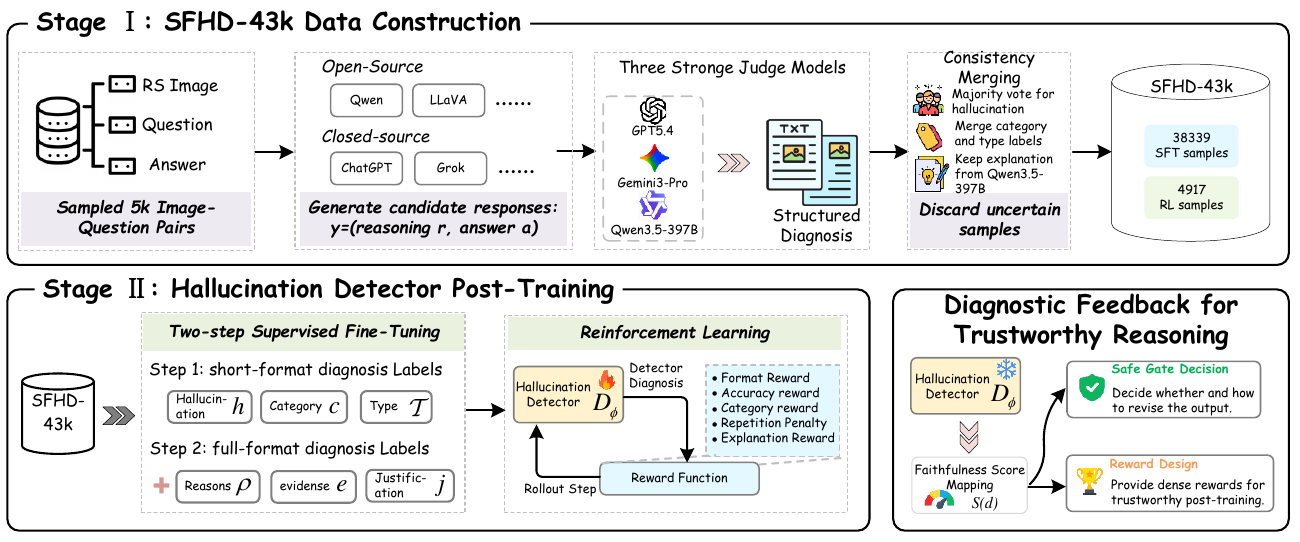}
    \caption{Overview of the structured fine-grained hallucination diagnosis pipeline.}
\label{fig:sfhd_pipeline}
\end{figure*}

{Among the many factors contributing to hallucinations~\cite{bai2024hallucination, huang2025survey}, a critical factor lies in reinforcement learning with verifiable rewards (RLVR), a widely adopted post-training paradigm that typically assesses only the final output holistically. Even when visual evidence is incorporated into the reward, such coarse-grained feedback offers no indication of the specific error span, the error type (e.g., object recognition, attribute perception, spatial reasoning, or logical deduction), or the appropriate correction strategy.}

{Addressing this limitation requires structured diagnosis that pinpoints where hallucinations occur, why they conflict with the visual evidence, and how they should be corrected, thereby converting coarse correctness signals into actionable guidance. A natural way to exploit such diagnosis is to let the model revise its own responses accordingly, in the spirit of self-reflection. However, correction itself is a double-edged sword: a response may arrive at the correct answer through unfaithful reasoning, in which case indiscriminately rewriting the entire response risks corrupting an answer that is already correct. Effective hallucination mitigation therefore demands not only fine-grained diagnosis, but also a correction mechanism that acts selectively on the flawed portion of a response.}

{Motivated by these observations, we propose FaithRS, a hallucination-aware self-reflection framework for multimodal reasoning in remote sensing, which converts structured hallucination diagnosis into dense supervisory signals for RLVR training. Specifically, we first construct SFHD-43K, a structured fine-grained hallucination diagnosis dataset built from diverse model outputs and consolidated through multi-model judge consensus, on which we train a detector that identifies hallucination categories and fine-grained types together with their causes, visual evidence, and diagnostic rationale. FaithRS then integrates these diagnostics into a detector-guided correction pipeline, where a safety gate determines the correction strategy based on hallucination status and answer correctness: reliable responses are retained, responses with both flawed reasoning and incorrect answers are fully rewritten, and responses with correct answers but unfaithful reasoning are revised while the answer is preserved—thereby realizing the selective correction discussed above and avoiding the answer degradation caused by indiscriminate rewriting. The correction trajectory is optimized with a composite reward covering answer accuracy, visual faithfulness, hallucination reduction, output validity, and correction safety, elevating hallucination diagnosis from a post-hoc evaluation signal to an explicit training constraint.}

{Experiments at the 0.8B, 4B, and 9B parameter scales on the Qwen3.5 backbone demonstrate consistent average gains across the MME-RealWorld-RS, RSVQA-HR, CHOICE, and LHRS benchmarks, with a maximum average improvement of 11.64 points; FaithRS-9B attains the best overall average performance among all evaluated models. Ablation studies show that detector-guided reinforcement learning and self-reflection contribute complementary gains, and training dynamics reveal that the model progressively improves the hallucination correction rate while maintaining a low answer degradation rate.}

{The main contributions are summarized as follows:}

\begin{itemize}
    \item We construct SFHD-43K and train a structured hallucination detector that identifies hallucination types, causes, and supporting visual evidence.
    
    \item We propose FaithRS, which uses detector-guided corrective rollout and safety-aware gating to learn whether and how to revise hallucinated reasoning.
    
    \item Experiments across three model scales and four remote sensing benchmarks show consistent average gains of up to 11.64 points, with ablations confirming the effectiveness of the proposed components.
\end{itemize}

% 2_formatting
\section{Methodology}
\subsection{Structured Fine-grained Hallucination Diagnosis}
To provide hallucination diagnostic signals for subsequent trustworthy reasoning post-training, as shown in Fig.~\ref{fig:sfhd_pipeline}, we first build a Structured Fine-grained Hallucination Detection Model. Unlike evaluators that return only scalar scores, the output of $D_\phi$ is constrained to output a parseable hallucination diagnosis structure. In this way, the detector can determine not only whether a model output contains hallucinations, but also the underlying source of each hallucination and its inconsistency with the visual evidence.

\begin{figure*}[t]
    \centering
    \includegraphics[width=0.98\linewidth]{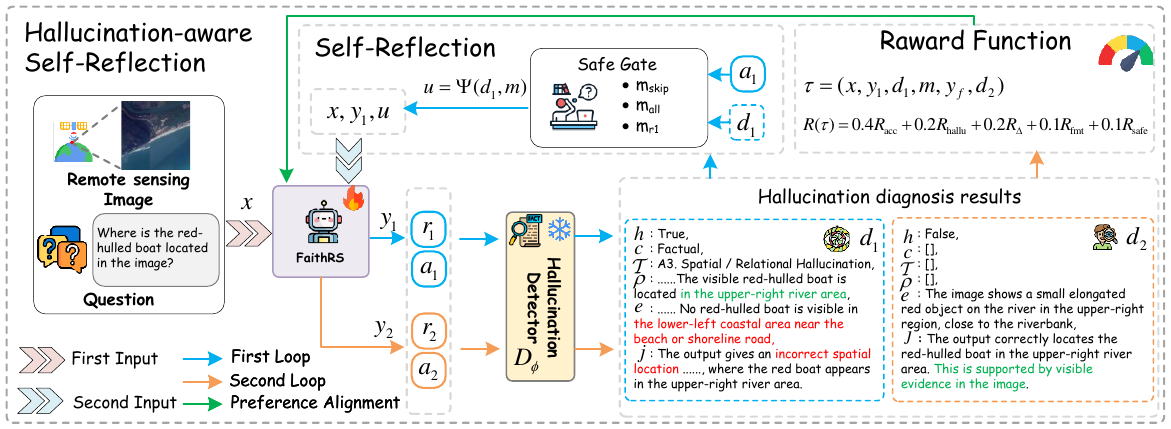}
    \caption{Overview of the hallucination-aware self-reflection training framework.}
\label{fig:FaithRS_pipeline}
\end{figure*}
\subsubsection{Stage 1: SFHD-43K Dataset Construction.}
\vspace{-0.2cm}
To train a detector capable of producing structured diagnostic results, we construct SFHD-43K, a structured fine-grained hallucination diagnosis dataset. The construction process consists of four steps: image-question sampling, candidate reasoning generation, multi-model judge hallucination annotation, and consistency merging. We first sample 5k image-question pairs from GeoLLaVA-8K \cite{wang2025geollava8k}. Each sample contains a remote sensing image, a natural-language question, and an answer, which provide the basic inputs for candidate reasoning generation and hallucination annotation. We then use multiple open-source and closed-source MLLMs to generate responses for the same input, forming a candidate reasoning set. Since different models exhibit different biases in visual perception, language reasoning, and spatial relation understanding, the resulting candidate set covers diverse output patterns. Compared with manually constructed error samples, this multi-model generation strategy more naturally captures typical failure modes of current MLLMs in remote sensing reasoning.

We then use three strong judge models to annotate hallucinations in the candidate outputs. Each judge receives the image, question, reference answer, model reasoning, and model answer, and produces a structured diagnosis $d=(h,c,\mathcal{T},\rho,e,j)$, where $h$ denotes whether a hallucination exists, $c$ denotes the hallucination category, $\mathcal{T}$ denotes the set of fine-grained hallucination types, $\rho$ describes the error causes, $e$ specifies the relevant visual evidence, and $j$ provides the diagnostic justification. We merge the hallucination indicator $h$ using strict majority voting. If no majority decision can be reached, the sample is assigned an uncertain label and removed from the training set. For the category $c$ and type set $\mathcal{T}$, we retain only the annotations from judges whose hallucination decisions agree with the majority result, and then perform order-preserving deduplication. For the explanatory components $\rho$, $e$, and $j$, we prioritize the output of Qwen3.5-397B-A17B \cite{qwen35blog} to improve the completeness and consistency of the error analysis and visual evidence description. After removing uncertain and invalid samples, we obtain SFHD-43K, which contains 38,339 samples for supervised fine-tuning (SFT) and 4,917 samples for RL.

\subsubsection{Stage 2: Training the Hallucination Detector.}
We use Qwen3.5-2B \cite{qwen35blog} as the backbone of the hallucination detector $D_\phi$ and train it through SFT followed by RL. The SFT stage consists of two phases. In the first phase, the detector is trained with short-form diagnostic targets $d_{\mathrm{short}}=(h,c,\mathcal{T})$. This phase enables the detector to learn hallucination classification, coarse-grained category prediction, and fine-grained type recognition, while establishing a stable and parseable output format. In the second phase, the supervision is extended to the full diagnosis $d=(h,c,\mathcal{T},\rho,e,j)$. This allows the detector to provide structured explanations and evidence-grounded judgments for subsequent corrective reasoning.

After SFT, we further optimize the detector with RL. The detector is trained with rewards for structural validity, hallucination decision accuracy, fine-grained label consistency, and output conciseness. Specifically, the format reward verifies whether the prediction forms a valid structured output with all required components. The hallucination reward measures whether the predicted decision $h$ agrees with the reference diagnosis. The type reward evaluates the consistency of the predicted category $c$ and type set $\mathcal{T}$ with their reference annotations. A repetition penalty discourages templated and redundant generations.

\subsubsection{Diagnostic Feedback for Trustworthy Reasoning.}
After two-stage SFT and RL optimization, $D_\phi$ can convert free-form model reasoning outputs into structured hallucination diagnostic feedback. This feedback is used in subsequent FaithRS training to guide the revision decisions of the Safe Gate and to provide parseable process-level signals for reward design. In the reward design section, we further define a deterministic mapping from the diagnostic result $d$ to the visual faithfulness score $S(d)$.

\subsection{Hallucination-aware Self-reflection Reasoning Model}
After obtaining the structured hallucination detector, we further train the Hallucination-aware Self-reflection Reasoning Model (FaithRS), whose overall training framework is illustrated in Fig.~\ref{fig:FaithRS_pipeline}. The goal is to enable the generation model to explicitly use detector feedback to correct reasoning content that is not grounded in visual evidence. Given an input $x=(I,Q)$, the model first generates an initial response. The hallucination detector $D_\phi$ then diagnoses the reasoning process in a structured form. If the diagnosis indicates hallucinations in the reasoning process, the model uses this feedback to generate a corrected response. In this way, hallucination feedback is no longer used only as a post-hoc evaluation signal. Instead, it becomes an explicit training condition that guides the model to revise its reasoning process under visual evidence constraints.

\subsubsection{Data Construction for FaithRS.}
The training corpus for FaithRS is composed of a supervised cold-start subset and a detector-guided RL subset. For the cold-start stage, we first construct an instruction-tuning dataset using a curated collection of remote sensing VQA data \cite{10056343} that covers both object-centric scenes and high-resolution aerial imagery. Specifically, we randomly sample 30k image, question, answer triples, then prompt GPT-4o to generate concise evidence-grounded reasoning for each item, normalising every sample into a reasoning, answer pair. This stage familiarises the model with the desired “think-then-answer” format and anchors the reasoning to observable visual cues, providing a stable template for the subsequent self-reflection RL phase.

For the RL stage, we construct a remote sensing visual question answering training pool from GeoLLaVA \cite{wang2025geollava8k}. We first collect about 20k image, question, answer triples and use the SFT-trained model to generate candidate responses for each sample. We then apply the trained hallucination detector $D_\phi$ to diagnose whether each response contains a hallucination. According to answer correctness and hallucination risk, the samples are divided into an easy training pool and a hard training pool. Easy samples are defined as those with correct initial answers and no detected hallucinations, whereas hard samples include those with incorrect initial answers or detected hallucination risks. This process yields 6,750 easy samples and 13,250 hard samples. We use the hard split for detector-guided self-reflection training, since these samples contain more complex errors and therefore require diagnostic feedback for effective correction.

\subsubsection{Hallucination-aware Self-reflection.}
Given an input $x=(I,Q)$, the reasoning model first generates an initial response:
\begin{equation}
y_1=(r_1,a_1)\sim \pi_\theta(\cdot\mid x),
\end{equation}
where $r_1$ denotes the initial reasoning process and $a_1$ denotes the initial answer. Given the reference answer $a^\ast $, we define the correctness of the initial answer as
\begin{equation}
c_1=C(a_1,a^\ast).
\end{equation}

The hallucination detector $D_\phi$ then performs structured diagnosis on the initial response:
\begin{equation}
d_1=D_\phi(I,Q,a^\ast,r_1,a_1).
\end{equation}

Based on the initial diagnosis $d_1$ and answer correctness $c_1$, FaithRS applies a Safe Gate to determine whether revision is required and which content should be revised:
\begin{equation}
m=\mathcal{G}(d_1,c_1)=
\begin{cases}
\mathsf{m}_{\mathrm{skip}}, & H(d_1)=0,\\
\mathsf{m}_{\mathrm{all}}, & H(d_1)=1,\ c_1=0,\\
\mathsf{m}_{\mathrm{r1}}, & H(d_1)=1,\ c_1=1,
\end{cases}
\end{equation}
where $H(d_1)\in{0,1}$ denotes whether the detector identifies hallucinated content, and $c_1\in{0,1}$ indicates whether the initial answer is correct. Under $\mathsf{m}_{\mathrm{skip}}$, the initial response is retained without an additional generation step. Under $\mathsf{m}_{\mathrm{all}}$, both the reasoning process and final answer are revised because the initial response contains hallucination and yields an incorrect answer. Under $\mathsf{m}_{\mathrm{r1}}$, only the reasoning process is revised while the correct answer is preserved. This routing mechanism avoids unnecessary revision when no hallucination is detected and prevents answer degradation when correcting an unfaithful reasoning chain.

When the routing function triggers revision, the structured diagnostic result $d_1$ is converted into a textual self-reflection signal:
\begin{equation}
u_1=\Psi(d_1,m),
\end{equation}
where $\Psi$ extracts the hallucination type, error cause, supporting visual evidence, and diagnostic rationale from the hallucination diagnosis, and organizes them into a revision prompt according to the routing mode. The model then generates a revised response conditioned on the input, the initial response, and the diagnostic feedback:
\begin{equation}
y_2=(r_2,a_2)\sim \pi_\theta(\cdot\mid x,y_1,u_1).
\end{equation}

We denote the final response as $y_f=(r_f,a_f)$, where $r_f$ and $a_f$ are the final self-reflective reasoning process and the final answer, respectively. For skipped samples, $y_f=y_1$; for revised samples, $y_f=y_2$. The final answer correctness is then computed, and the hallucination detector is invoked again to obtain the final diagnosis:
\begin{equation}
c_2=C(a_f,a^\ast),\qquad
d_2=D_\phi(I,Q,a^\ast,r_f,a_f).
\end{equation}

Each training trajectory therefore contains both the initial state $(y_1,d_1,c_1)$ and the final state $(y_f,d_2,c_2)$. This paired structure allows the training objective to evaluate not only the quality of the final response, but also whether the model actually corrects hallucinated reasoning in the initial response.

The policy model is optimized with a trajectory-level composite reward. For each two-stage training trajectory $\tau=(x,y_1,d_1,m,y_f,d_2)$, we construct the reward from five aspects: final answer correctness $R_{\mathrm{acc}}$, final hallucination faithfulness $R_{\mathrm{hallu}}$, hallucination improvement before and after revision $R_{\Delta}$, output format $R_{\mathrm{fmt}}$, and revision safety $R_{\mathrm{safe}}$:
\begin{equation}
\begin{aligned}
R(\tau)=&\,0.40R_{\mathrm{acc}}
+0.20R_{\mathrm{hallu}}+0.20R_{\Delta}\\
&+0.10R_{\mathrm{fmt}}+0.10R_{\mathrm{safe}}.
\end{aligned}
\end{equation}

The answer accuracy reward is given by the final answer correctness:
\begin{equation}
R_{\mathrm{acc}}=c_2=C(a_f,a^\ast).
\end{equation}

The hallucination-aware reward is mapped from the final diagnostic result $d_2$:
\begin{equation}
R_{\mathrm{hallu}}=S(d_2),
\end{equation}
where $S(\cdot)\in[0,1]$ denotes the hallucination-aware faithfulness score (Appendix~\ref{app:faithfulness_score}) determined by the structured diagnostic result. A higher score indicates more faithful reasoning and a lower degree of hallucination. The hallucination improvement reward explicitly measures the change in faithfulness before and after revision. Let $s_1=S(d_1)$, $s_2=S(d_2)$, and $\Delta s=s_2-s_1$. We use a monotonic piecewise function to convert $\Delta s$ into an improvement reward:
\begin{equation}
R_{\Delta}=\Gamma(\Delta s),
\end{equation}
where $\Gamma(\cdot)$ assigns higher rewards to substantial improvements and penalizes faithfulness degradation (Appendix~\ref{app:delta_reward}). This encourages the model not only to produce a high-quality final response, but also to repair hallucinated reasoning in the initial response. The format reward $R_{\mathrm{fmt}}$ constrains the model to produce a concise and parseable reasoning-answer structure. The safety reward $R_{\mathrm{safe}}$ penalizes answer degradation caused by revision. In particular, under the $\mathsf{m}_{\mathrm{r1}}$ mode, if the initial answer is reliable but the revised answer loses correctness, the safety reward is set to a low value. This prevents the model from damaging an originally correct answer while repairing the reasoning process.

For the same input $x_i$, the policy model samples $G$ two-stage trajectories $\{\tau_{i,g}\}_{g=1}^{G}$ and computes the composite reward for each trajectory:
\begin{equation}
R_{i,g}=R(\tau_{i,g}).
\end{equation}

The rewards are then normalized within the candidate trajectory group of the same input to obtain the relative advantage:
\begin{equation}
\hat{A}_{i,g}=
\frac{
R_{i,g}-\operatorname{mean}_{g'}(R_{i,g'})
}{
\operatorname{std}_{g'}(R_{i,g'})+\varepsilon
}.
\end{equation}

This within-group normalization makes the optimization focus on the relative quality of different candidate trajectories under the same image-question input, rather than on absolute reward-scale differences across samples.

We optimize FaithRS with GSPO \cite{zheng2025gspo}. Unlike methods based on token-level \cite{shao2024deepseekmath} importance ratios, GSPO computes the likelihood ratio between the new and old policies at the sequence level. For the final response $y_{f,i,g}$ of the $g$-th trajectory, we denote its context as $\operatorname{ctx}_{i,g}$. If the sample does not trigger revision, then $\operatorname{ctx}_{i,g}=x_i$ and $y_{f,i,g}=y_{1,i,g}$. If the sample triggers revision, then $\operatorname{ctx}_{i,g}=(x_i,y_{1,i,g},u_{1,i,g})$ and $y_{f,i,g}=y_{2,i,g}$. For brevity, define $p_{\eta,i,g,t}=\pi_\eta(y_{f,i,g,t}\mid \operatorname{ctx}_{i,g},y_{f,i,g,<t})$ for $\eta\in\{\theta,\theta_{\mathrm{old}}\}$. The sequence-level importance ratio is defined as
\begin{equation}
\begin{aligned}
w_{i,g}^{\mathrm{GSPO}}(\theta)
&=\exp\left(
\frac{1}{|y_{f,i,g}|}
\sum_{t=1}^{|y_{f,i,g}|}
\log\frac{p_{\theta,i,g,t}}{p_{\theta_{\mathrm{old}},i,g,t}}
\right).
\end{aligned}
\end{equation}
Based on this sequence-level ratio, the clipped GSPO objective is
\begin{equation}
\begin{aligned}
\mathcal{J}_{\mathrm{GSPO}}(\theta)
&=\mathbb{E}_{i,g}\bigl[
\min\bigl(
w_{i,g}^{\mathrm{GSPO}}(\theta)\hat{A}_{i,g},\\
&\qquad
\operatorname{clip}\left(
w_{i,g}^{\mathrm{GSPO}}(\theta),
1-\epsilon_c,
1+\epsilon_c
\right)\hat{A}_{i,g}
\bigr)\bigr].
\end{aligned}
\end{equation}
where $\epsilon_c$ is the clipping coefficient, and $\theta_{\mathrm{old}}$ denotes the old policy parameters used to sample the trajectories.

% 3_finalcopy
\section{Experiments}
\begin{table*}[t]
\centering

\resizebox{0.93\textwidth}{!}{
\begin{tabular}{llccccc}
\toprule
Model & Publication & MME-RealWorld-RS & RSVQA-HR & CHOICE & LHRS & Avg \\
\midrule
\multicolumn{7}{l}{\textit{General Domain MLLMs}} \\
GPT-4o\cite{hurst2024gpt}
& arXiv 2024& 21.36& 60.82& 65.66& 66.15& 53.50 \\
Gemini-2.5-Pro\cite{comanici2025gemini}
& arXiv 2025& 38.06& 67.49& 72.48& 66.41& 61.11 \\
Claude-Haiku-4.5
& --& 21.83& 38.79& 59.82& 54.03& 43.62 \\
Qwen3.5-0.8B~\cite{qwen35blog}
& --& 38.36& 53.48& 51.90& 45.40& 47.29 \\
Qwen3.5-4B~\cite{qwen35blog}
& --& 53.22& 65.92& 70.84& 68.54& 64.63 \\
Qwen3.5-9B~\cite{qwen35blog}
& --& \best{62.09}& 67.97& \second{75.15}& \third{70.54}& \third{68.94} \\
LLaVA-v1.6-Vicuna-7B~\cite{liu2023visual}
& NeurIPS 2023& 38.87& --& 53.32& 52.87& -- \\
LLaMA3-LLaVA-8B~\cite{chen2024open}
& --& 47.57& --& 63.47& 55.50& -- \\
InternVL3.5-8B~\cite{wang2025internvl3}
& arXiv 2025& 52.00& 73.83& 62.70& 54.52& 60.76 \\
DeepSeek-VL-7B-Chat~\cite{lu2024deepseekvl}
& arXiv 2024& 34.96& 67.54& 62.07& 47.98& 53.14 \\
\midrule
\multicolumn{7}{l}{\textit{Remote sensing specialized MLLMs}} \\
GeoChat~\cite{kuckreja2023geochat}
& CVPR 2024& 29.24& 66.70& 52.89& 44.09& 48.23 \\
VHM~\cite{pang2024vhm}
& AAAI 2025& 38.38& 75.93& 58.35& 49.42& 55.52 \\
EarthDial~\cite{soni2024earthdial}
& CVPR 2025& 33.88& 70.91& 59.75& 49.30& 53.46 \\
GeoZero~\cite{wang2025geozero}
& arXiv 2025& 51.44& \second{79.03}& \third{74.81}& 64.45& 67.43 \\
RemoteReasoner~\cite{yao2026remotereasoner}
& AAAI 2026& 30.29& 66.77& 69.02& 57.18& 55.82 \\
RSThinker~\cite{liu2025rsthinker}
& ICLR 2026& 38.41& 63.52& 56.70& 58.35& 54.25 \\
\addlinespace[2pt]
FaithRS-0.8B
& Ours& 47.00& \third{78.23}& 57.17& 53.30& 58.93 \\
FaithRS-4B
& Ours& \third{57.33}& 77.55& 72.72& \best{74.52}& \second{70.53} \\
FaithRS-9B
& Ours& \second{57.51}& \best{81.85}& \best{76.86}& \second{72.71}& \best{72.23} \\
\bottomrule
\end{tabular}
}
\caption{Quantitative comparison of general-purpose and remote sensing MLLMs on four remote sensing benchmarks. Avg denotes the arithmetic mean across MME-RealWorld-RS, RSVQA-HR, CHOICE, and LHRS. Dark, medium, and light green indicate the best, second-best, and third-best results among models with reported scores in each column, respectively.}
\label{tab:main_results}
\end{table*}

\begin{figure*}[t]
\centering
\includegraphics[width=0.94\linewidth]{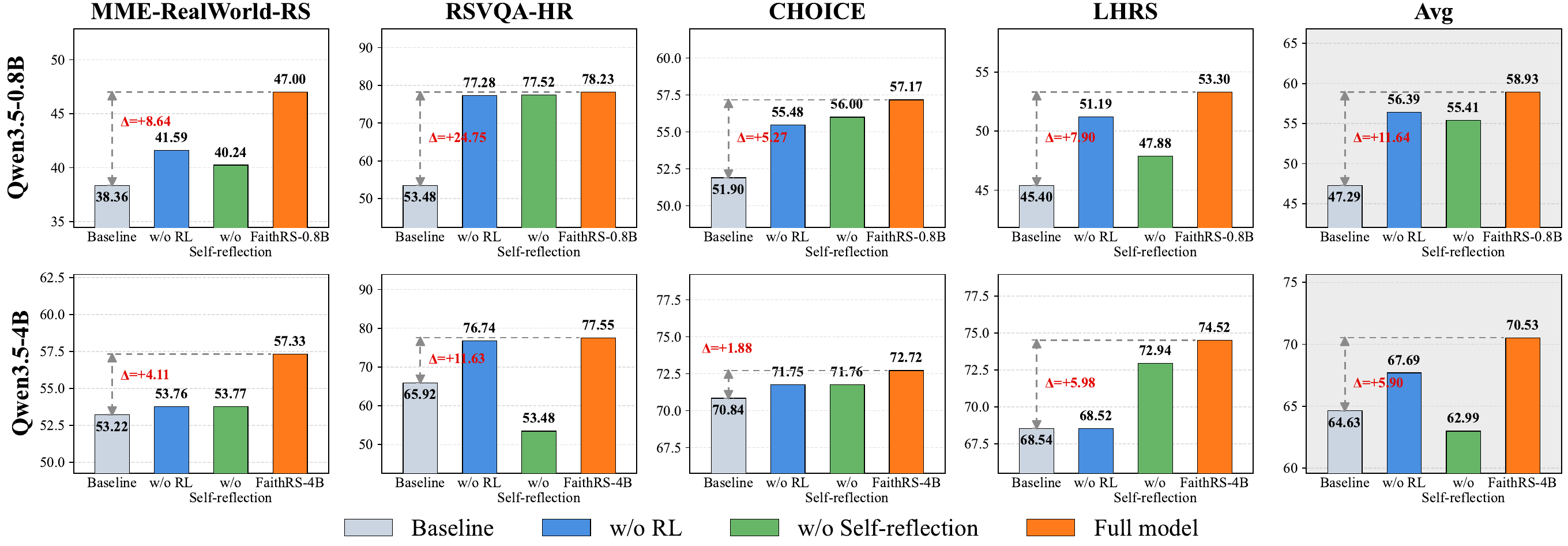}
\caption{Ablation study of FaithRS at the 0.8B and 4B scales.}
\label{fig:ablation_results}
\end{figure*}

\subsection{Experimental Setup} 
\subsubsection{Implementation Details.} We train FaithRS using Qwen3.5 backbone models at three scales, namely 0.8B, 4B, and 9B \cite{qwen35blog}. Before reinforcement learning, we conduct supervised cold-start training with LoRA \cite{hu2022lora} applied to all linear layers. The learning rate is set to $2\times10^{-4}$, and the maximum sequence length is set to 4096, allowing the model to learn a stable reasoning-answer output format. During reinforcement learning, we optimize the model with GSPO \cite{zheng2025gspo}. For each input question, the model samples 8 rollouts. The per-device batch size is set to 4, the model is trained for 1 epoch, and the learning rate is set to $1\times10^{-5}$. The sampling temperature is set to 0.8. All training experiments are conducted on 4 A100 GPUs with 80GB memory.

\subsubsection{Evaluation Benchmarks.} We evaluate task performance on four remote sensing multimodal understanding benchmarks, including MME-RealWorld-RS \cite{zhang2024mmerealworld}, RSVQA-HR \cite{lobry2020rsvqa}, CHOICE \cite{an2025choice}, and LHRS \cite{muhtar2024lhrsbot}. These benchmarks cover low- and high-resolution remote sensing scene understanding and reasoning in real-world scenarios, providing a broad evaluation of remote sensing perception, question answering, and reasoning abilities. 

\subsubsection{Baselines.} To comprehensively evaluate the effectiveness of FaithRS, we compare it with two groups of baselines, following the categorization in Table~\ref{tab:main_results}. The first group consists of general-purpose MLLMs, including the closed-source models GPT-4o \cite{hurst2024gpt}, Gemini-2.5-Pro \cite{comanici2025gemini}, and Claude-Haiku-4.5, as well as the open-source models Qwen3.5-0.8B, Qwen3.5-4B, Qwen3.5-9B \cite{qwen35blog}, LLaVA-v1.6-Vicuna-7B \cite{liu2023visual}, LLaMA3-LLaVA-8B  \cite{chen2024open}, InternVL3.5-8B \cite{wang2025internvl3}, and DeepSeek-VL-7B-Chat \cite{lu2024deepseekvl}. The second group consists of remote sensing MLLMs, including GeoChat \cite{kuckreja2023geochat}, VHM \cite{pang2024vhm}, EarthDial \cite{soni2024earthdial}, GeoZero \cite{wang2025geozero}, RemoteReasoner \cite{yao2026remotereasoner}, and RSThinker \cite{liu2025rsthinker}. We re-evaluate all available baseline models using their official checkpoints under identical prompts, decoding configurations, and evaluation scripts.

\subsection{Main Results}
Table~\ref{tab:main_results} reports the quantitative results on four remote sensing multimodal understanding benchmarks. Overall, FaithRS consistently improves the average performance of all three Qwen3.5 backbones. FaithRS-0.8B achieves an average score of 58.93, improving upon Qwen3.5-0.8B by 11.64 points. FaithRS-4B increases the average score from 64.63 to 70.53, yielding a gain of 5.90 points. FaithRS-9B further reaches 72.23, outperforming its backbone by 3.29 points and achieving the best average result among all evaluated models. These improvements across 0.8B, 4B, and 9B backbones demonstrate that the proposed hallucination-aware self-reflection training framework is effective across different model capacities.

Compared with existing remote sensing specialized MLLMs, FaithRS exhibits a clear overall advantage. FaithRS-4B achieves an average score of 70.53, surpassing GeoZero by 3.10 points despite using a smaller backbone than many competing models. FaithRS-9B further improves the average score to 72.23 and establishes the strongest overall performance in the table. Even FaithRS-0.8B reaches 58.93, outperforming GeoChat, VHM, EarthDial, RemoteReasoner, and RSThinker in terms of average score. These results indicate that remote sensing instruction tuning or task-specific adaptation alone does not fully exploit the reasoning capacity of MLLMs. By incorporating structured fine-grained hallucination diagnosis and detector-guided self-reflection into post-training, FaithRS achieves stronger and more balanced performance across remote sensing perception, question answering, and reasoning tasks.

Although FaithRS-9B improves substantially on RSVQA-HR, CHOICE, and LHRS, its MME-RealWorld-RS score decreases from 62.09 to 57.51. This suggests that the benefits of hallucination-aware self-reflection are task dependent and may involve a trade-off between conservative evidence-grounded generation and performance on certain real-world evaluation formats. Nevertheless, the consistent improvements in average performance across all three model scales, together with the strong results on RSVQA-HR, CHOICE, and LHRS, confirm the overall effectiveness and scalability of FaithRS.

\begin{figure*}[ht]
    \centering
    \includegraphics[width=0.95\linewidth]{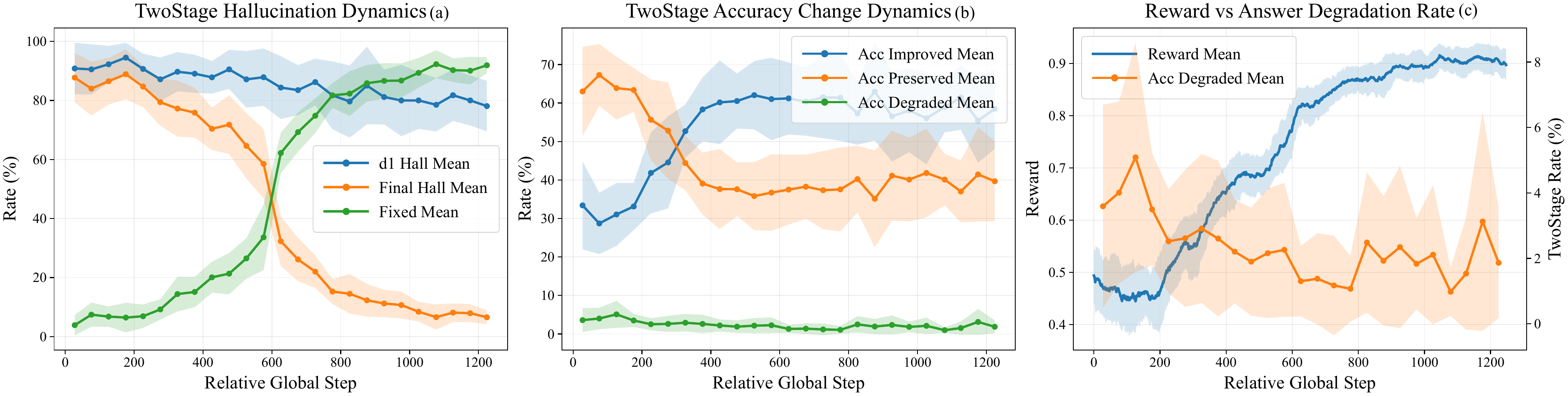}
    \caption{Training dynamics of FaithRS-0.8B during detector-guided corrective rollout. 
    }
\label{fig:training_dynamics}
\end{figure*}

\subsection{Ablation Study}

To examine the contribution of the main post-training components, we conduct ablation studies on FaithRS-0.8B and FaithRS-4B. We evaluate a variant without RL, denoted as w/o RL, and a variant without the self-reflection mechanism, denoted as w/o Self-reflection. Figure~\ref{fig:ablation_results} compares these variants with their corresponding Qwen3.5 backbones and the FaithRS models on four remote sensing benchmarks and the average score.

At the 0.8B scale, both ablated variants improve over the original backbone, but neither matches the full model. FaithRS-0.8B achieves an average score of 58.93, compared with 56.39 for w/o RL and 55.41 for w/o Self-reflection, and consistently performs best across all four benchmarks. A similar trend appears at the 4B scale, where the full model reaches 70.53, while removing reinforcement learning or self-reflection reduces the average score to 67.69 and 62.99, respectively. The larger drops on MME-RealWorld-RS, RSVQA-HR, and LHRS indicate that detector-guided reinforcement learning and self-reflection are important for complex visual reasoning and corrective generation. Overall, supervised cold-start training provides stable reasoning and answer generation, while detector-guided reinforcement learning and self-reflection enable the model to use diagnostic feedback for targeted correction. Their combination yields the strongest and most balanced performance.

\subsection{Training Dynamics Analysis}
To examine how FaithRS acquires self-reflective revision ability, we analyze the training dynamics of FaithRS-0.8B during Two-Stage Corrective Rollout. Figure~\ref{fig:training_dynamics}(a) compares the hallucination rates of the initial and final responses. The initial hallucination rate remains high because the RL training set consists of challenging samples that require correction. As training proceeds, the hallucination rate of the final responses decreases steadily, while the hallucination repair rate rises substantially. This trend indicates that the model progressively learns to use structured detector feedback to identify unreliable content and produce revisions that are better grounded in visual evidence. Figure~\ref{fig:training_dynamics}(b) presents the answer change patterns before and after revision. Answer preservation and answer improvement account for most revisions, whereas answer degradation remains consistently rare. This behavior is consistent with the intended routing strategy: the model preserves reliable answers while correcting unsupported reasoning, and performs broader revisions only when both the reasoning and answer are unreliable. FaithRS therefore learns a selective, feedback-conditioned revision policy rather than an indiscriminate rewriting strategy. Figure~\ref{fig:training_dynamics}(c) shows the relationship between the training reward and answer degradation. Although the reward increases throughout training, the answer degradation rate remains low and exhibits no corresponding upward trend. This result suggests that FaithRS does not improve its reward by sacrificing answer correctness. Instead, the Safe Gate and safety reward constrain the revision process and reduce the risk of damaging initially reliable answers.

Overall, these training dynamics show that FaithRS gradually develops the ability to reduce hallucinations without compromising answer reliability. Structured diagnosis identifies what should be corrected, Two-Stage Corrective Rollout turns this feedback into learnable revision behavior, and the safety mechanism controls the risk introduced by revision.

\section{Conclusion}
This work investigates visually unfaithful reasoning in remote sensing MLLMs and proposes FaithRS, a hallucination-aware self-reflection framework for faithful remote sensing reasoning. FaithRS first employs a structured fine-grained hallucination detector to identify error types, causes, and supporting visual evidence. It then incorporates the diagnostic feedback into Safe Detector-Guided Corrective Rollout, where a safety-aware gate determines whether and how the initial response should be revised. A composite reward further encourages answer accuracy, visual faithfulness, effective hallucination correction, valid output format, and revision safety. Unlike conventional single-stage reinforcement learning based only on terminal feedback, FaithRS explicitly learns a diagnostic-feedback-conditioned revision policy while concentrating optimization on the final response. Experiments across multiple remote sensing benchmarks show that FaithRS consistently improves different Qwen3.5 backbones, reduces factual and logical hallucinations, and maintains a low answer degradation rate during revision. These results demonstrate the potential of diagnosis-guided self-reflection to improve the reliability of remote sensing multimodal reasoning through structured diagnosis, conditional revision, and safety-aware optimization.

\bibliographystyle{plainnat}
\bibliography{references}

\clearpage
\appendix
\part*{Appendices}
\addcontentsline{toc}{part}{Appendices}

% Appendix: related_work
\section{Additional Related Work}

\subsection{Multimodal Reasoning Models for Remote Sensing}

In recent years, MLLMs have been increasingly applied to remote sensing image understanding. Early studies typically adapt general-domain MLLMs to the remote sensing domain through supervised fine-tuning on remote sensing image--text pairs, visual question answering samples, and instruction-following data. These adaptations enable models to describe remote sensing scenes, recognize objects, and interact with users through natural-language instructions. Representative methods, including RSGPT, GeoChat, SkySenseGPT, RS-GPT4V, SkyEyeGPT, and LHRS-Bot, construct remote sensing instruction datasets and employ vision--language alignment strategies to improve performance in scene classification, object localization, visual question answering, and relation understanding \citeapp{app_hu2023rsgpt, app_kuckreja2023geochat, app_luo2024skysensegpt, app_xu2024rsgpt4v, app_zhan2024skyeyegpt, app_muhtar2024lhrsbot}. Other studies incorporate stronger visual encoders, region-level visual representations, or preference optimization to improve perception and response generation in complex remote sensing scenes \citeapp{app_soni2024earthdial, app_pang2024vhm, app_wang2025geollava8k}. Collectively, these studies demonstrate the potential of MLLMs for open-ended remote sensing understanding.

Recent studies have begun to investigate reliability evaluation and hallucination detection for remote sensing MLLMs, aiming to determine whether model responses are consistent with the evidence contained in remote sensing images \citeapp{app_pang2024vhm, app_liu2026rshbench}. However, most existing studies focus primarily on evaluation or detection. Although they can identify the presence of hallucinations, they rarely transform diagnostic information into learnable supervision for model training. For remote sensing multimodal reasoning, determining whether a response contains hallucinations is insufficient. A reliable model must also identify whether an error arises from object recognition, attribute description, spatial relation judgment, or logical inference, while preserving reliable content during revision. In contrast to existing methods, this work explicitly integrates fine-grained hallucination diagnosis into the training process. The Hallucination Detector produces structured diagnostic feedback, enabling the reasoning model to learn a more faithful self-correction policy under constraints imposed by remote sensing visual evidence.

\subsection{Multimodal Hallucinations and Visual Faithfulness}

Multimodal hallucinations generally refer to inconsistencies between generated textual responses and the provided visual content \citeapp{app_li2023pope, app_liu2026rshbench}. Existing studies indicate that hallucinations in MLLMs arise from the combined effects of training data, model architecture, and optimization objectives. At the data level, large-scale image--text corpora are often collected from the web or filtered using heuristic rules, which can introduce noisy annotations, weakly aligned samples, and statistical biases. Instruction-tuning data frequently relies on language-model-generated content, which may further introduce descriptions unsupported by visual evidence. At the model level, the perceptual capability of the visual encoder, the representation capacity of the cross-modal alignment interface, and the knowledge priors encoded in the language model jointly affect how visual evidence is incorporated into generation. At the training level, MLLMs generally inherit the autoregressive next-token prediction objective, which optimizes the likelihood of text generation but imposes no explicit consistency constraints on visual facts such as objects, attributes, and spatial relations. These limitations are further amplified in remote sensing scenes because of small object scales, ambiguous land-cover boundaries, dense spatial layouts, and complex topological relations. Consequently, models are more likely to generate content that conflicts with visual evidence regarding object existence, attribute descriptions, and spatial relations \citeapp{app_kuckreja2023geochat, app_wang2025geollava8k, app_liu2026rshbench}.

% Appendix: reward_details
\section{Complete Reward Definitions}
\subsection{Hallucination-Aware Faithfulness Score \texorpdfstring{$S(d)$}{S(d)}}
\label{app:faithfulness_score}

The structured detector diagnosis is $d=(h,c,\mathcal{T},\rho,e,j)$. Here, $h\in\{0,1\}$ indicates whether a hallucination is detected, $c$ denotes the set of hallucination categories, $\mathcal{T}$ denotes the set of fine-grained hallucination types, $\rho$ describes the error causes, $e$ specifies the relevant visual evidence, and $j$ provides the diagnostic justification. Only $h$, $c$, and $\mathcal{T}$ contribute to the scalar faithfulness score. The explanatory fields $\rho$, $e$, and $j$ are used to construct revision feedback and support qualitative auditing.

The final two-stage reward uses the following visual faithfulness score:
\begin{equation}
S(d)=
\begin{cases}
0.50, & \substack{d\text{ is missing}\\
                   \text{or cannot be parsed}},\\
1.00, & h=0,\\
\begin{aligned}
&0.25\,r_{\mathrm{bin}}(h)+0.25\,r_{\mathrm{cat}}(c)\\
&\qquad +0.50\,r_{\mathrm{sub}}(\mathcal{T})
\end{aligned}
& h=1.
\end{cases}
\label{eq:faithfulness_score}
\end{equation}
Here, $r_{\mathrm{bin}}(h)=0$ for a hallucinated response. The category-level score is defined as
\begin{equation}
r_{\mathrm{cat}}(c)=
\begin{cases}
0, & Factual\,Hallucination\in c,\\
0.3, & Logical\,Hallucination\in c.
\end{cases}
\label{eq:category_score}
\end{equation}

To compute the subtype-level score, we first assign each fine-grained hallucination type a severity value:
\begin{equation}
\begin{aligned}
\mu(\mathrm{A1})&=\mu(\mathrm{A2})=\mu(\mathrm{A3})=1.00,\\
\mu(\mathrm{B1})&=\mu(\mathrm{B3})=0.40,\\
\mu(\mathrm{B2})&=0.30,\qquad
\mu(\mathrm{B4})=0.20.
\end{aligned}
\label{eq:subtype_severity}
\end{equation}
Unknown or missing subtype labels are assigned a default severity of $0.52$. For a diagnosis containing multiple subtypes, we use the maximum subtype severity and apply an additional multiplicity penalty:
\begin{equation}
p(\mathcal{T})
=
\min\left(0.10,\,
0.03\max\bigl(|\mathcal{T}|-1,0\bigr)\right).
\label{eq:multiplicity_penalty}
\end{equation}
The subtype-level faithfulness score is then
\begin{equation}
r_{\mathrm{sub}}(\mathcal{T})
=
\operatorname{clip}_{[0,1]}
\left(
1-\max_{t\in\mathcal{T}}\mu(t)-p(\mathcal{T})
\right),
\label{eq:subtype_score}
\end{equation}
where the maximum severity is set to $0.52$ when $\mathcal{T}$ is empty. Therefore, more severe or more numerous hallucination types result in a lower faithfulness score.

\subsection{Delta Reward \texorpdfstring{$R_{\Delta}$}{R-delta}}
\label{app:delta_reward}
Let $d_k=(h_k,c_k,\mathcal{T}_k,\rho_k,e_k,j_k)$, for $k\in\{1,2\}$, denote the detector diagnoses of the initial and final responses, respectively. We define the change in the hallucination score as
\begin{equation}
\Delta s = S(d_2)-S(d_1).
\end{equation}
The delta reward is computed using the following piecewise function:
\begin{equation}
\Gamma(\Delta s)=
\begin{cases}
1.00, & \Delta s \ge 0.4,\\
0.80, & 0.2 \le \Delta s < 0.4,\\
0.65, & 10^{-6} < \Delta s < 0.2,\\
0.50, & |\Delta s| \le 10^{-6},\\
0.25, & -0.2 < \Delta s < -10^{-6},\\
0.00, & \Delta s \le -0.2.
\end{cases}
\end{equation}
Accordingly, the reward wrapper returns $R_{\Delta}=\Gamma(\Delta s)$ when both cached hallucination scores are available. Otherwise, it assigns the neutral reward $R_{\Delta}=0.5$.

\subsection{Safe Reward \texorpdfstring{$R_{\mathrm{safe}}$}{R-safe}}
Let $a_1$ and $a_2$ be the initial and revised answer correctness scores. The safe reward is
\begin{equation}
R_{\mathrm{safe}}=
\begin{cases}
1.00, & a_2 > a_1 + 10^{-6},\\
0.70, & a_2 \ge a_1 - 10^{-6},\\
0.00, & \text{otherwise}.
\end{cases}
\end{equation}
In the mode $\mathsf{m}_{\mathrm{r1}}$, if $a_1\ge \tau$ and the revised answer degrades by more than the epsilon tolerance, the reward is forced to $0$. This additional rule protects answer-correct samples from being rewritten into worse answers while the reasoning is revised.

\subsection{GSPO Objective}
For a prompt group indexed by $i$ and a sampled completion indexed by $g$, the code uses group normalization:
\begin{equation}
\hat A_{i,g}=\frac{R_{i,g}-\operatorname{mean}(\{R_{i,j}\}_{j=1}^{G})}{\operatorname{std}(\{R_{i,j}\}_{j=1}^{G})+\epsilon}.
\end{equation}
The sequence-level importance ratio is computed from the average log-ratio across completion tokens:
\begin{equation}
w_{i,g}^{\mathrm{GSPO}}(\theta)
=
\left(
\frac{\pi_\theta(y_{i,g}\mid x_i)}{\pi_{\theta_{\mathrm{old}}}(y_{i,g}\mid x_i)}
\right)^{\frac{1}{|y_{i,g}|}}.
\end{equation}
The trainer then applies asymmetric clipping with lower and upper bounds $1-\epsilon_{\mathrm{low}}$ and $1+\epsilon_{\mathrm{high}}$, where the script sets $\epsilon_{\mathrm{low}}=3\times 10^{-4}$ and $\epsilon_{\mathrm{high}}=4\times 10^{-4}$. The optimization is final-response-only because the loss scale is set to \texttt{last\_round}.

% Appendix: prompts
\newtcblisting{promptbox}[1]{
  enhanced,
  breakable,
  listing only,
  listing engine=listings,
  colback=white,
  colframe=black,
  colbacktitle=white,
  coltitle=black,
  fonttitle=\bfseries,
  title={#1},
  sharp corners,
  boxrule=0.5pt,
  left=2mm,
  right=2mm,
  top=1.5mm,
  bottom=1.5mm,
  before skip=0.8em,
  after skip=1em,
  listing options={
    style=promptstyle,
    frame=none,
    xleftmargin=0pt,
    xrightmargin=0pt
  }
}

\section{Complete Prompt Templates}

\subsection{FaithRS Initial Generation Prompt}
The generator SFT and RL prompts share a common remote sensing VQA instruction:
\begin{promptbox}{Initial Generation Prompt}
You are a remote sensing visual question answering expert.
Answer based solely on what you directly observe in the image.
Use <think>...</think> for your reasoning, then write ONLY the final answer after </think>.

Answer format rules:
- Yes/No questions: answer 'yes' or 'no'
- Count questions: answer with a number only
- Other questions: answer in 1-5 words
\end{promptbox}
The RL data append \texttt{/nothink} to the user turn so that the generator remains in the raw-answer data format used by the training scripts.

\subsection{Revision Prompts}
The two revision prompts used by the Safe Gate are:
\begin{promptbox}{Revision Prompt: Reasoning and Answer}
The previous answer contains hallucination and may be incorrect.

Detector feedback:
- Types: {types}
- Problems: {reasons}
- Evidence / missing evidence: {evidence}
- Justification: {justification}

You must:
1. Remove unsupported claims.
2. Correct wrong object, attribute, count, or spatial claims.
3. Revise both reasoning and answer if needed.
4. Use only visible evidence from the image.

Output strict JSON:
{
  "reasoning": "...",
  "answer": "..."
}
\end{promptbox}

\begin{promptbox}{Revision Prompt: Reasoning Only}
The previous final answer matches the ground truth, but the reasoning contains unsupported claims.

Detector feedback:
- Types: {types}
- Problems: {reasons}
- Evidence / missing evidence: {evidence}
- Justification: {justification}

You must:
1. Preserve the final answer unless the image clearly contradicts it.
2. Remove unsupported reasoning.
3. Replace hallucinated claims with visually grounded evidence.
4. Output strict JSON only:
{
  "reasoning": "...",
  "answer": "..."
}
\end{promptbox}

% Appendix: additional_analysis
\section{Additional Training Analysis}

This section first examines which structured diagnostic fields and training stages are important for the hallucination detector, and then analyzes the optimization behavior of FaithRS using the accompanying training diagnostics. In the curve analysis, all horizontal axes report relative global steps, solid curves denote the reported means, and shaded regions denote the corresponding standard deviations. Because several rates are computed over different event subsets, we interpret their trajectories separately rather than assuming that all curves share the same denominator.

\subsection{Hallucination Detector Ablations}

We evaluate the hallucination detector from two complementary perspectives. First, we compare detector performance across the short-form SFT, full-form SFT, and RL stages to examine how structured prediction quality evolves throughout training. Second, we compare the baseline RL configuration with configurations involving different combinations of the explanatory fields $\rho$, $e$, and $j$, which correspond to \texttt{hallucination\_reasons}, \texttt{evidence\_basis}, and \texttt{justification}, respectively. Higher values indicate better performance for all reported metrics.

\paragraph{Training-stage comparison.} Table~\ref{tab:detector_stage_ablation} compares detector performance after short-form SFT, full-form SFT, and RL optimization. Moving from short-form to full-form SFT increases the parse rate by $34.60$ percentage points, accuracy by $0.1406$, F1 by $0.2629$, and MCC by $0.2304$. Subsequent RL optimization further improves the parse rate by $41.50$ percentage points, accuracy by $0.2128$, F1 by $0.2880$, and MCC by $0.2995$. These results show a consistent improvement throughout the training pipeline. Full-form supervision provides richer structured targets and substantially improves both output validity and classification performance, while RL further strengthens format stability and diagnostic accuracy. Because these stages are applied sequentially, the results should be interpreted as a comparison of training progression rather than as a strictly controlled component ablation.

\begin{table}[tb]
\centering
\small
\setlength{\tabcolsep}{4pt}
\renewcommand{\arraystretch}{1.12}
\begin{tabular}{lcccc}
\toprule
Stage & Parse (\%) & Acc. & F1 & MCC \\
\midrule
SFT-Short & 22.30 & 0.5331 & 0.3497 & 0.2386 \\
SFT-Full & 56.90 & 0.6737 & 0.6126 & 0.4690 \\
RL (Best) & \textbf{98.40} & \textbf{0.8865} & \textbf{0.9006} & \textbf{0.7685} \\
\bottomrule
\end{tabular}
\caption{Comparison across hallucination detector training stages. Bold values indicate the best result in each column.}
\label{tab:detector_stage_ablation}
\end{table}

\paragraph{Diagnostic-field ablation during RL.} Table~\ref{tab:detector_field_ablation} compares the baseline RL configuration with configurations involving different combinations of the explanatory fields. The Baseline achieves the strongest overall performance, including the best parse rate, accuracy, precision, F1, MCC, and Factual-F1. Specifically, it reaches a parse rate of $98.40\%$, an overall F1 of $0.9006$, and an MCC of $0.7685$. The configuration involving $e$ achieves the highest recall of $0.9649$ and the highest Logical-F1 of $0.7385$, but its precision and MCC decrease to $0.8050$ and $0.6959$, respectively. Similar recall-oriented shifts are observed in several other field configurations, indicating a trade-off between sensitivity and balanced classification performance.

\begin{table*}[tb]
\centering
\small
\setlength{\tabcolsep}{3.2pt}
\renewcommand{\arraystretch}{1.12}
\begin{tabular}{lcccccccc}
\toprule
Setting & Parse (\%) & Acc. & Prec. & Recall & F1 & MCC & Factual-F1 & Logical-F1 \\
\midrule
Baseline
  & \textbf{98.40} & \textbf{0.8865} & \textbf{0.9014} & 0.8998
  & \textbf{0.9006} & \textbf{0.7685} & \textbf{0.9033} & 0.7221 \\
w/ $\rho$
  & 91.50 & 0.8534 & 0.8784 & 0.8629
  & 0.8706 & 0.7018 & 0.8672 & 0.7322 \\
w/ $e$
  & 97.70 & 0.8464 & 0.8050 & \textbf{0.9649}
  & 0.8777 & 0.6959 & 0.8794 & \textbf{0.7385} \\
w/ $j$
  & 95.30 & 0.8484 & 0.8472 & 0.8963
  & 0.8711 & 0.6891 & 0.8338 & 0.6391 \\
w/ $\rho{+}e$
  & 96.60 & 0.8685 & 0.8911 & 0.8770
  & 0.8840 & 0.7323 & 0.8623 & 0.7060 \\
w/ $\rho{+}j$
  & 92.30 & 0.8404 & 0.8024 & 0.9561
  & 0.8725 & 0.6818 & 0.8760 & 0.7348 \\
w/ $e{+}j$
  & 96.80 & 0.8243 & 0.7830 & 0.9578
  & 0.8617 & 0.6517 & 0.8616 & 0.7051 \\
w/ $\rho{+}e{+}j$
  & 96.70 & 0.8544 & 0.8897 & 0.8506
  & 0.8697 & 0.7059 & 0.8569 & 0.6648 \\
\bottomrule
\end{tabular}
\caption{Ablation of structured diagnostic fields during detector RL. Bold values indicate the best result in each column.}
\label{tab:detector_field_ablation}
\end{table*}

The field-level results further show that parseability and classification quality do not necessarily improve together. Several configurations retain parse rates above $96\%$ but exhibit lower F1 and MCC than the Baseline. The largest MCC reduction is $0.1168$ for the $e{+}j$ configuration. This indicates that producing structurally valid outputs is not sufficient to guarantee balanced discrimination between hallucinated and non-hallucinated responses or between factual and logical hallucinations.

Overall, the training-stage comparison shows that full-form SFT and subsequent RL progressively improve structured prediction quality. However, the diagnostic-field configurations do not outperform the Baseline in terms of overall balanced performance. Although some configurations improve recall or Logical-F1, these gains are accompanied by reductions in precision, F1, or MCC. Since the tables report point estimates without confidence intervals, the observed differences establish the empirical ordering of the evaluated settings but should not be interpreted as evidence of statistical significance.

\subsection{Detector-Guided RL Training Dynamics}

\begin{figure}[tb]
  \centering
  \includegraphics[width=0.97\linewidth]{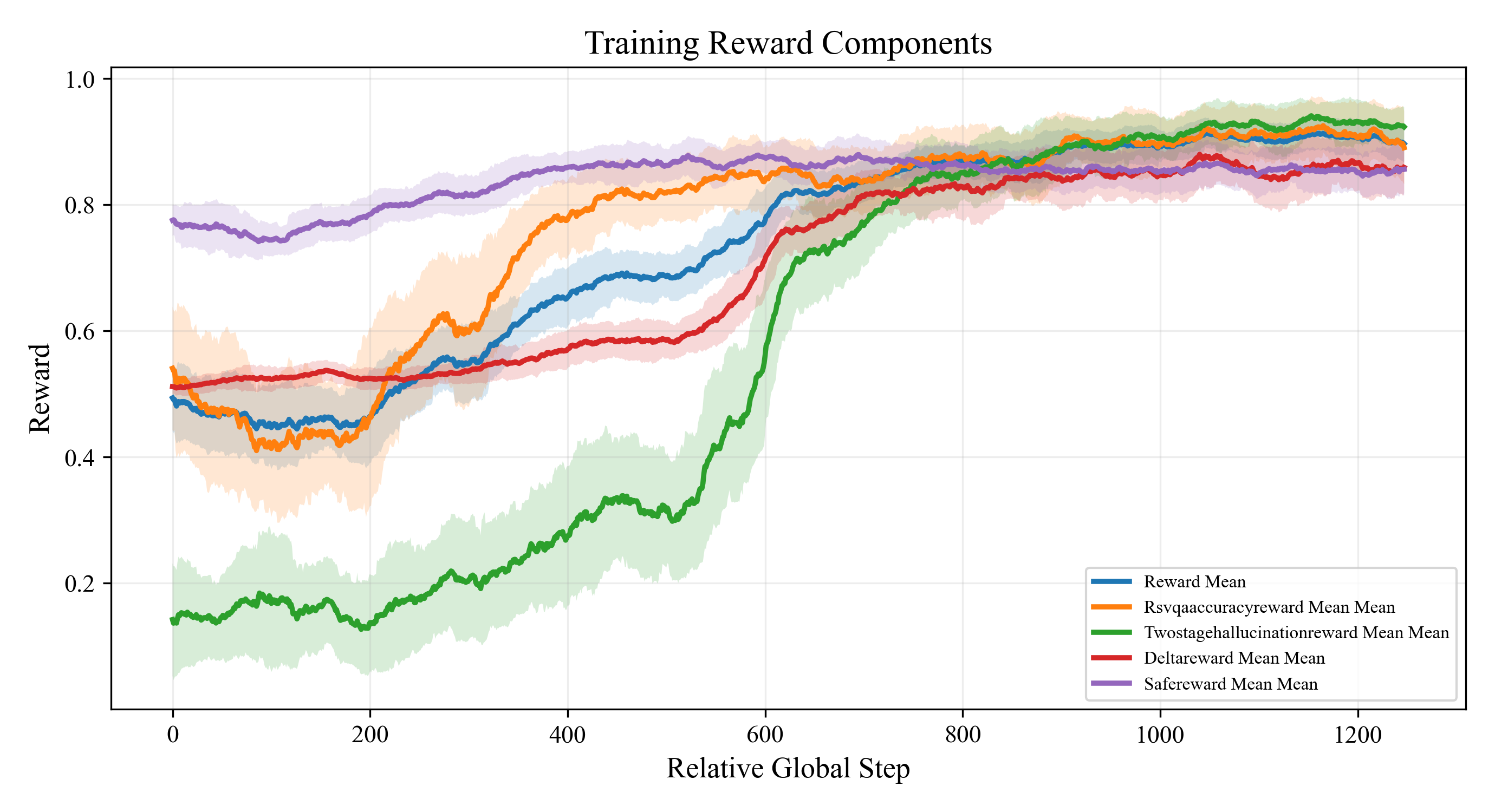}
  \caption{Evolution of the composite reward and its constituent terms. Solid curves show means and shaded regions show standard deviations.}
  \label{fig:reward_components}
\end{figure}

The training diagnostics reveal a coherent two-phase optimization process with a common transition around relative step 600. As shown in Figure~\ref{fig:reward_components}, the mean composite reward first decreases slightly from approximately $0.49$ to $0.45$ during the first 200 steps, then rises to about $0.82$ near step 600 and stabilizes around $0.89$--$0.91$ after step 900. The early decrease coincides with a temporary reduction in the answer-accuracy reward, whereas the subsequent improvement is shared across all reward components.

\begin{figure*}[tb]
  \centering
  \begin{minipage}[t]{0.32\textwidth}
    \centering
    {\small\bfseries Hallucination outcomes\par}
    \includegraphics[width=\linewidth]{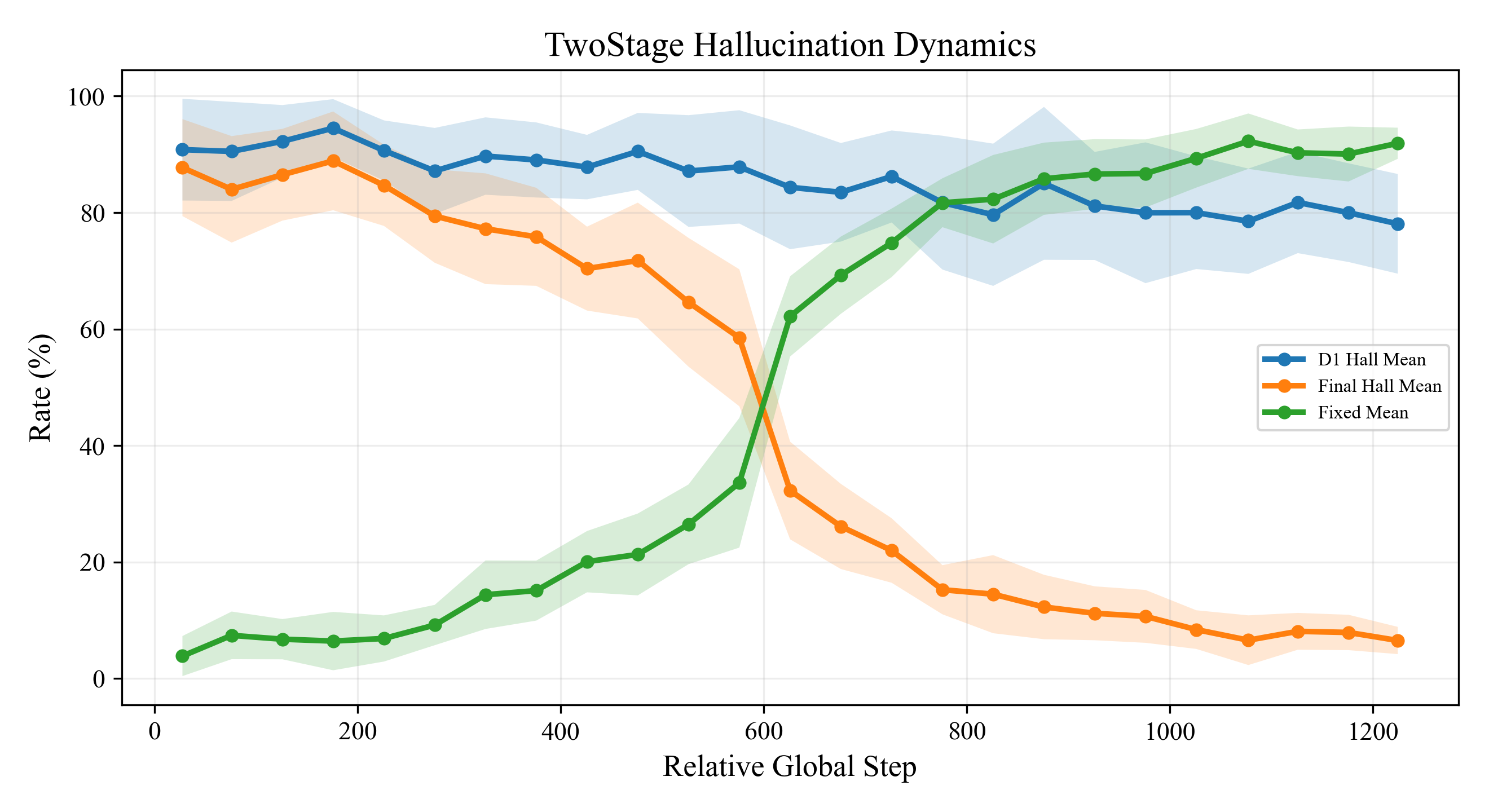}
  \end{minipage}
  \hfill
  \begin{minipage}[t]{0.32\textwidth}
    \centering
    {\small\bfseries Safe-Gate routing\par}
    \includegraphics[width=\linewidth]{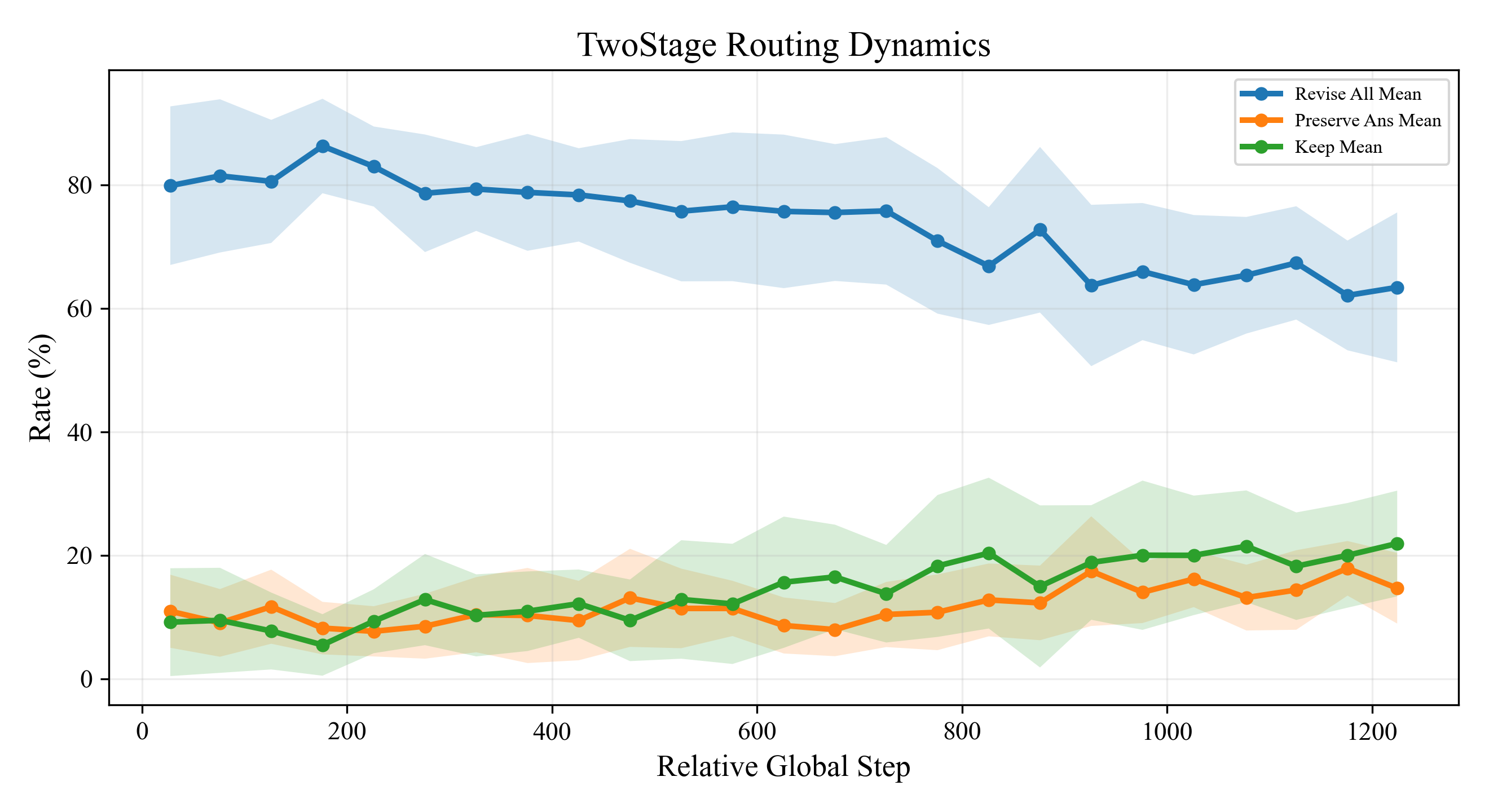}
  \end{minipage}
  \hfill
  \begin{minipage}[t]{0.32\textwidth}
    \centering
    {\small\bfseries Answer-correctness outcomes\par}
    \includegraphics[width=\linewidth]{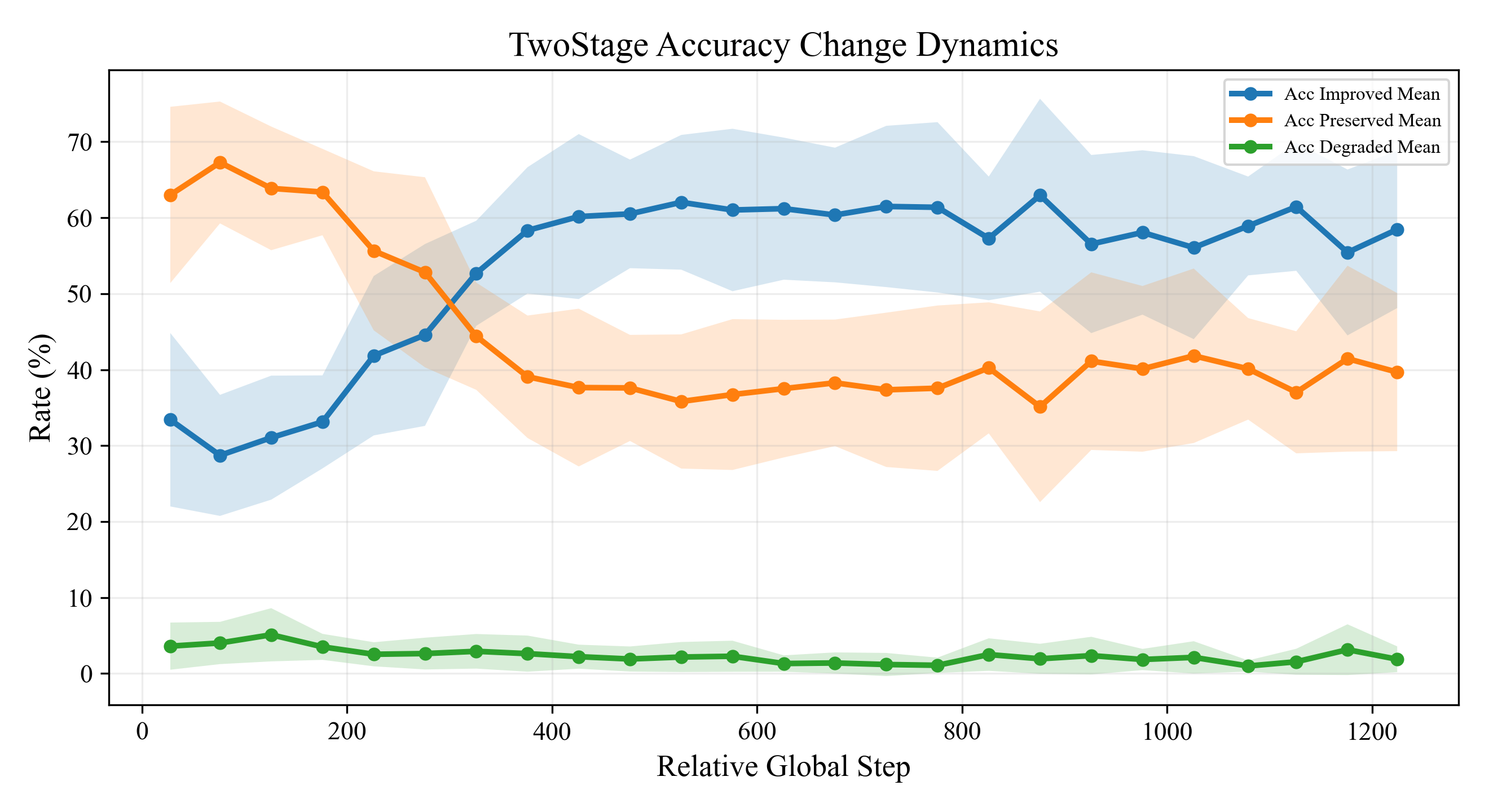}
  \end{minipage}
  \caption{Core detector-guided training dynamics. Left: initial hallucination, final hallucination, and fix rates. Middle: Safe-Gate routing rates. Right: improved, preserved, and degraded answer-correctness outcomes.}
  \label{fig:appendix_training_dynamics}
\end{figure*}

The hallucination reward contributes the largest relative change, increasing from roughly $0.14$ at the beginning of training to above $0.90$ near convergence, with its sharpest rise between approximately steps 550 and 650. Over the same period, the answer-accuracy reward recovers from an early minimum near $0.42$ to approximately $0.90$, and the delta reward increases from about $0.50$ to $0.86$. In contrast, the safe reward remains comparatively stable within an approximately $0.75$--$0.87$ range. Thus, the improvement in hallucination-sensitive rewards is not accompanied by a collapse of the answer-preservation objective. The generally narrower uncertainty bands near convergence further indicate more stable late-stage behavior.

The two detector passes further localize the source of this improvement. In the left panel of Figure~\ref{fig:appendix_training_dynamics}, the initial-diagnosis hallucination rate (\emph{D1 Hall}) remains high throughout training and decreases only moderately from approximately $90\%$ to around $80\%$, which is consistent with the use of the hard detector-guided RL subset. By contrast, the reported final hallucination rate falls from roughly $85\%$--$90\%$ to below $10\%$, while the fix rate rises from below $10\%$ to above $90\%$. The largest separation between the initial and final diagnoses appears around steps 580--650, matching the transition in the reward curves. This pattern is consistent with the policy learning to use detector feedback on difficult first-pass responses rather than obtaining a higher reward merely by avoiding revision.

As correction becomes more reliable, the Safe Gate also adopts a more selective routing distribution. In the middle panel of Figure~\ref{fig:appendix_training_dynamics}, the broad \emph{revise-all} route decreases from approximately $80\%$ to the mid-$60\%$ range, while the \emph{keep} route grows from below $10\%$ to around $20\%$ and the \emph{preserve-answer} route increases from roughly $8\%$--$10\%$ to approximately $15\%$. The policy therefore does not improve by sending an increasing fraction of samples through unrestricted rewriting. Instead, the higher reward coincides with more frequent use of routes that retain reliable content or restrict the scope of revision.

The accuracy transitions provide a final safety check on this routing behavior. The right panel of Figure~\ref{fig:appendix_training_dynamics} shows that the proportion of revisions that improve answer correctness rises from approximately $30\%$ during early training to around $58\%$--$62\%$ after step 400. The preserved-correctness rate decreases from roughly $65\%$ to approximately $35\%$--$42\%$ as more samples move into the improved category, whereas the degraded fraction remains small and declines from an early range of approximately $3\%$--$5\%$ to mostly $1\%$--$3\%$ during later training.

Taken together, the detector ablations and generator diagnostics support a coherent training account. Complete structured targets and RL provide the strongest observed combination of detector parseability and diagnostic quality, while detector-guided generator training aligns higher reward with more successful repair, more selective routing, and infrequent answer degradation. These results characterize the evaluated training process and do not replace held-out evaluation. In particular, they do not establish causal effects, formal statistical significance, or generalization beyond the evaluated settings. Moreover, the plotted rates may use different conditioning sets and should not be combined arithmetically; the supported evidence lies in their individual trajectories and aligned transition points.

% Appendix: qualitative_cases_selected
\section{Qualitative Examples on MME-RealWorld-RS}

We present nine representative examples from MME-RealWorld-RS covering color recognition, object counting, and spatial localization. For each question type, the selected cases include stable successes shared by all three FaithRS scales, scale-dependent corrections, and representative reasoning failures. The consistently correct examples demonstrate that FaithRS preserves reliable visual question answering behavior across different model capacities, while the failure cases reveal limitations that are not fully reflected by aggregate accuracy. 

\begin{figure*}[p]
\centering
\includegraphics[width=0.93\linewidth,height=0.27\textheight,keepaspectratio]{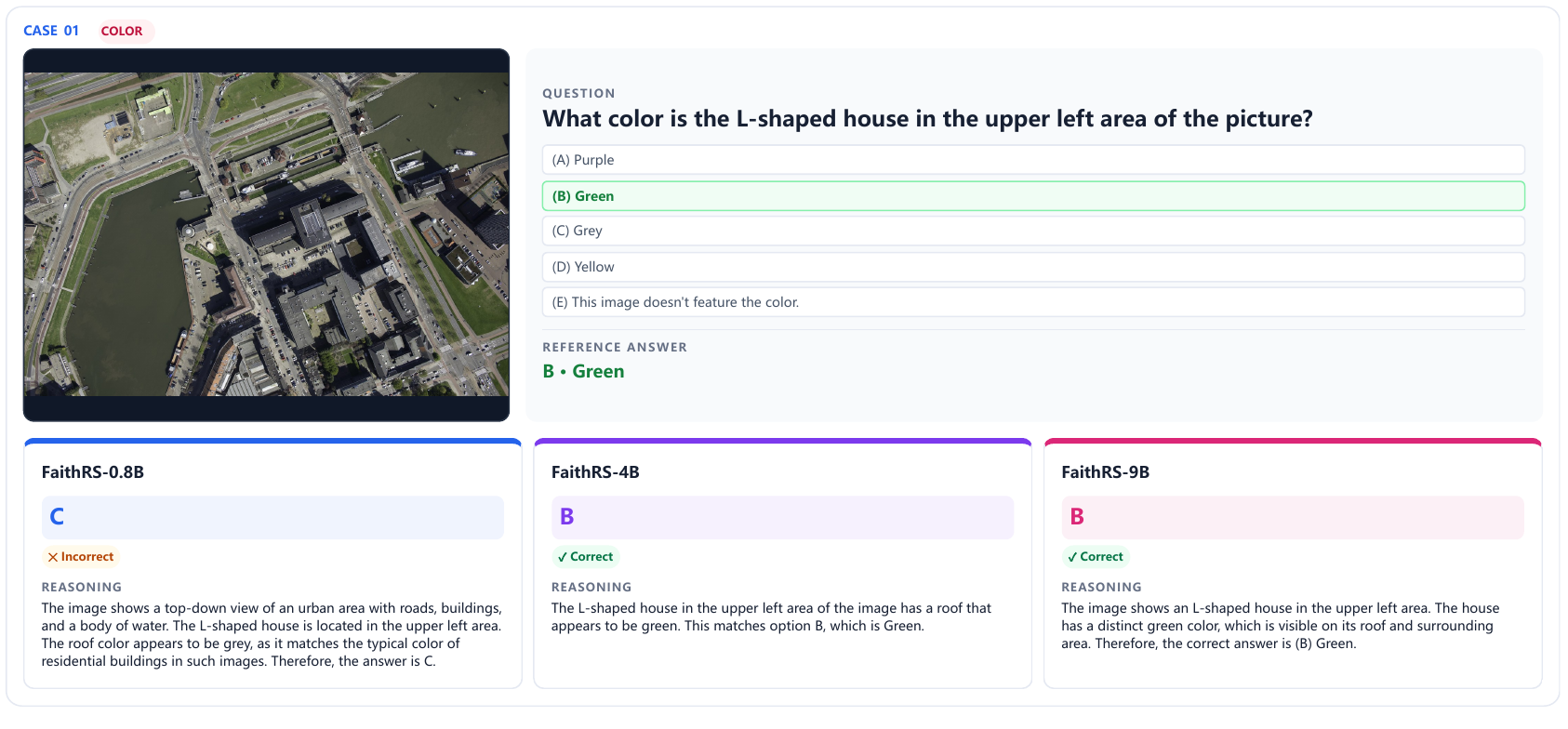}
\includegraphics[width=0.93\linewidth,height=0.27\textheight,keepaspectratio]{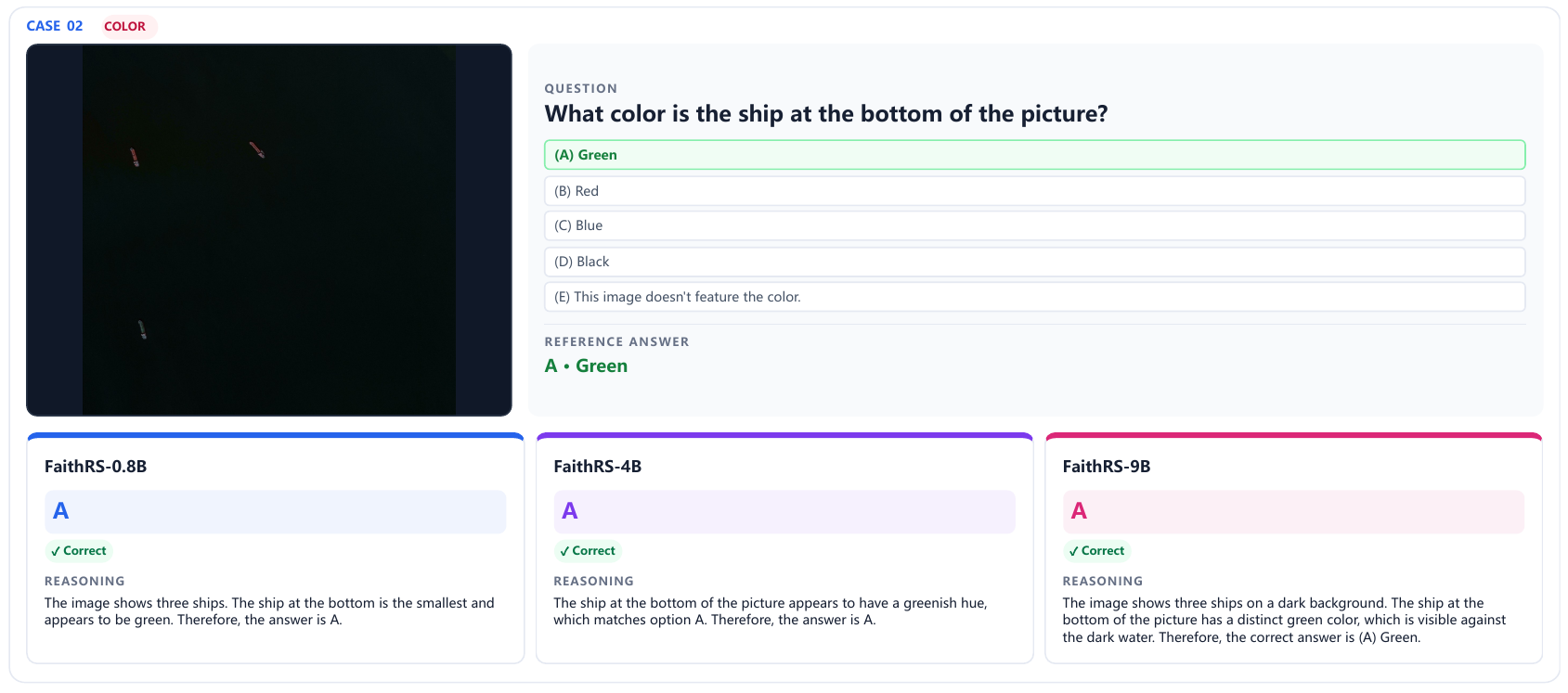}
\includegraphics[width=0.93\linewidth,height=0.27\textheight,keepaspectratio]{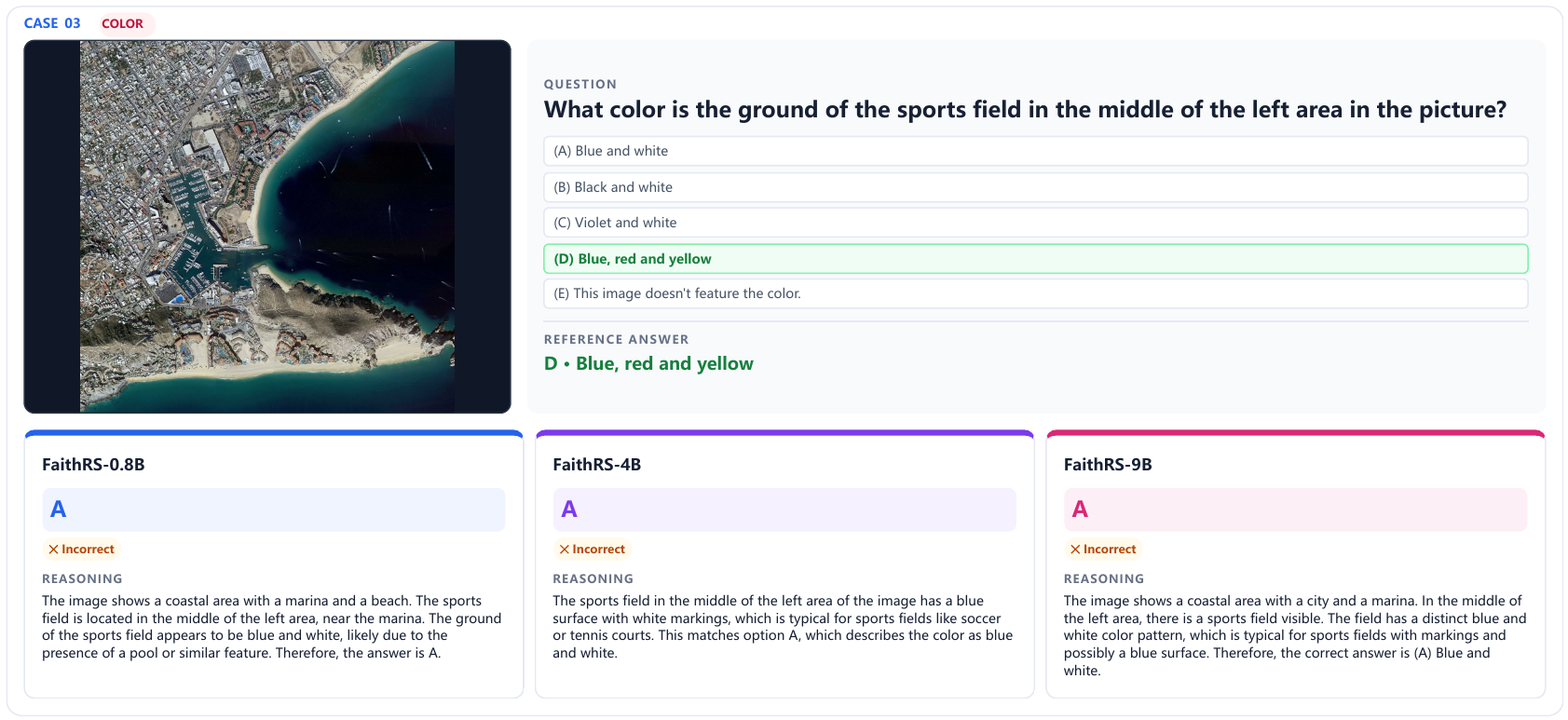}
\caption{Representative color-recognition examples from MME-RealWorld-RS. Case 1 shows a scale-dependent correction of a prior-driven attribute error. Case 2 is answered correctly by all three FaithRS scales, demonstrating stable color recognition. Case 3 remains a shared multicolor-recognition failure across all three models.}
\label{fig:qualitative_color}
\end{figure*}

\paragraph{Color recognition.} Figure~\ref{fig:qualitative_color} presents a scale-dependent correction, a stable success, and a shared failure. In Case 1, FaithRS-0.8B classifies the L-shaped house as gray and explains that gray is a typical color for residential buildings. This explanation relies partly on a generic scene prior rather than on the observed target roof. FaithRS-4B and FaithRS-9B instead identify the roof as green, consistent with the reference answer. Case 2 provides a stable positive example: all three models correctly identify the ship at the bottom of the image as green and support the answer with concise and mutually consistent visual descriptions. This result shows that the target attribute can be reliably preserved across model scales when the object is sufficiently distinguishable. By contrast, Case 3 remains challenging for all three models. Each model describes the sports field as blue and white, while the reference answer specifies blue, red, and yellow. The explanations further rationalize the incomplete prediction by referring to colors that are typical of sports fields or pool-like structures. This shared error indicates that increasing model capacity alone does not fully resolve fine-grained multicolor recognition when the target occupies only a small region of a wide remote sensing scene.

\begin{figure*}[p]
\centering
\includegraphics[width=0.93\linewidth,height=0.27\textheight,keepaspectratio]{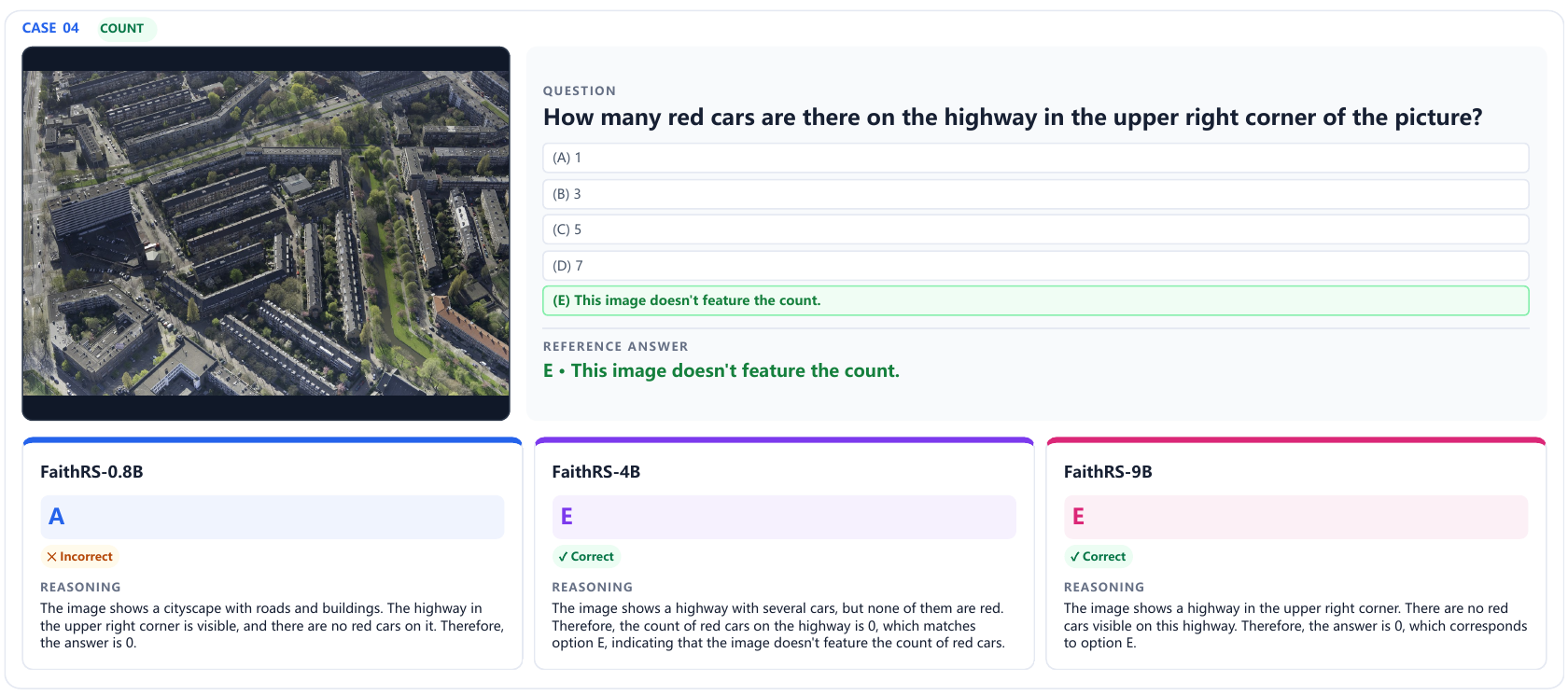}
\includegraphics[width=0.93\linewidth,height=0.27\textheight,keepaspectratio]{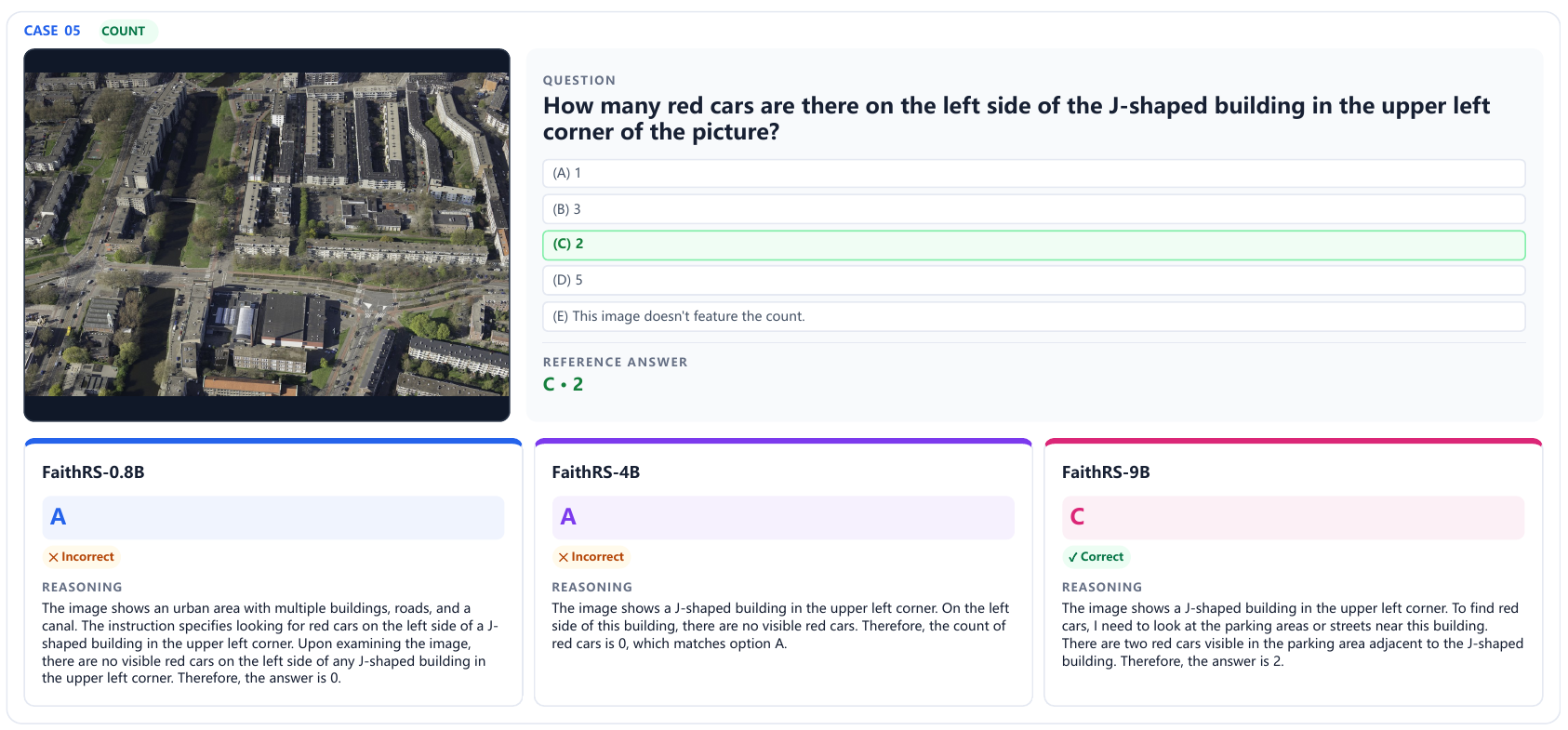}
\includegraphics[width=0.93\linewidth,height=0.27\textheight,keepaspectratio]{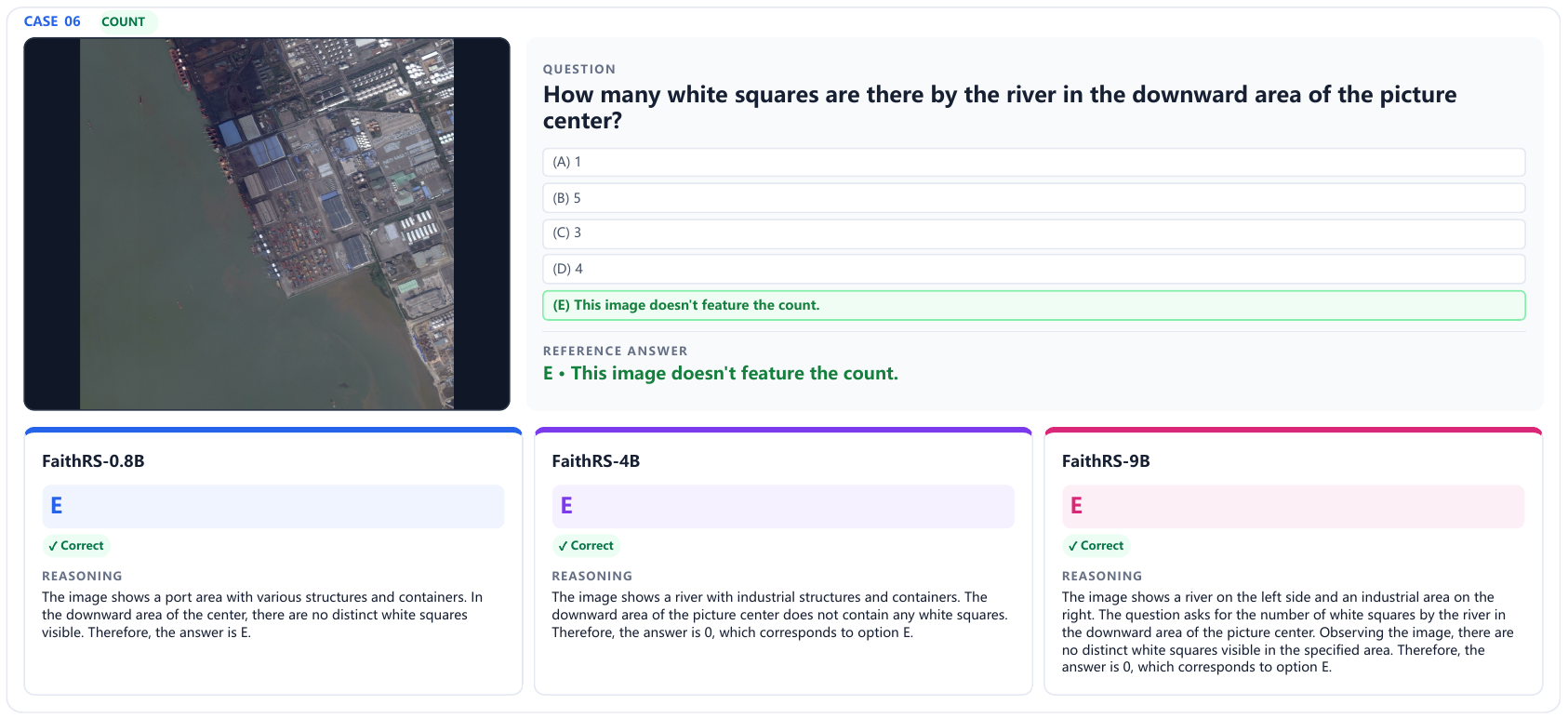}
\caption{Representative counting examples from MME-RealWorld-RS. Case 4 exposes an inconsistency between the absence judgment in the reasoning and the final option selected by FaithRS-0.8B. Case 5 shows a scale-dependent small-target omission, with only FaithRS-9B correctly identifying the two red cars. Case 6 is answered correctly by all three models and demonstrates stable target-absence recognition.}
\label{fig:qualitative_count}
\end{figure*}

\paragraph{Counting and target absence.} Figure~\ref{fig:qualitative_count} illustrates three distinct counting behaviors. In Case 4, FaithRS-0.8B explicitly states that no red cars are visible on the highway, but selects option A, which denotes one red car. The visual conclusion in the reasoning is therefore inconsistent with the final multiple-choice answer. FaithRS-4B and FaithRS-9B correctly map the absence judgment to option E. Case 5 reflects a different failure mode. FaithRS-0.8B and FaithRS-4B fail to detect the two red cars beside the J-shaped building, whereas FaithRS-9B correctly identifies both targets. This result suggests that small-object counting in dense urban scenes remains sensitive to model capacity. Case 6 provides a stable absence-detection example: all three models correctly determine that no white squares are present in the specified region and consistently select option E. The contrast between Cases 5 and 6 shows that selecting the no-target option is not inherently difficult. The main challenge is distinguishing genuine target absence from the presence of visually small or weakly contrasted objects.

\begin{figure*}[p]
\centering
\includegraphics[width=0.93\linewidth,height=0.27\textheight,keepaspectratio]{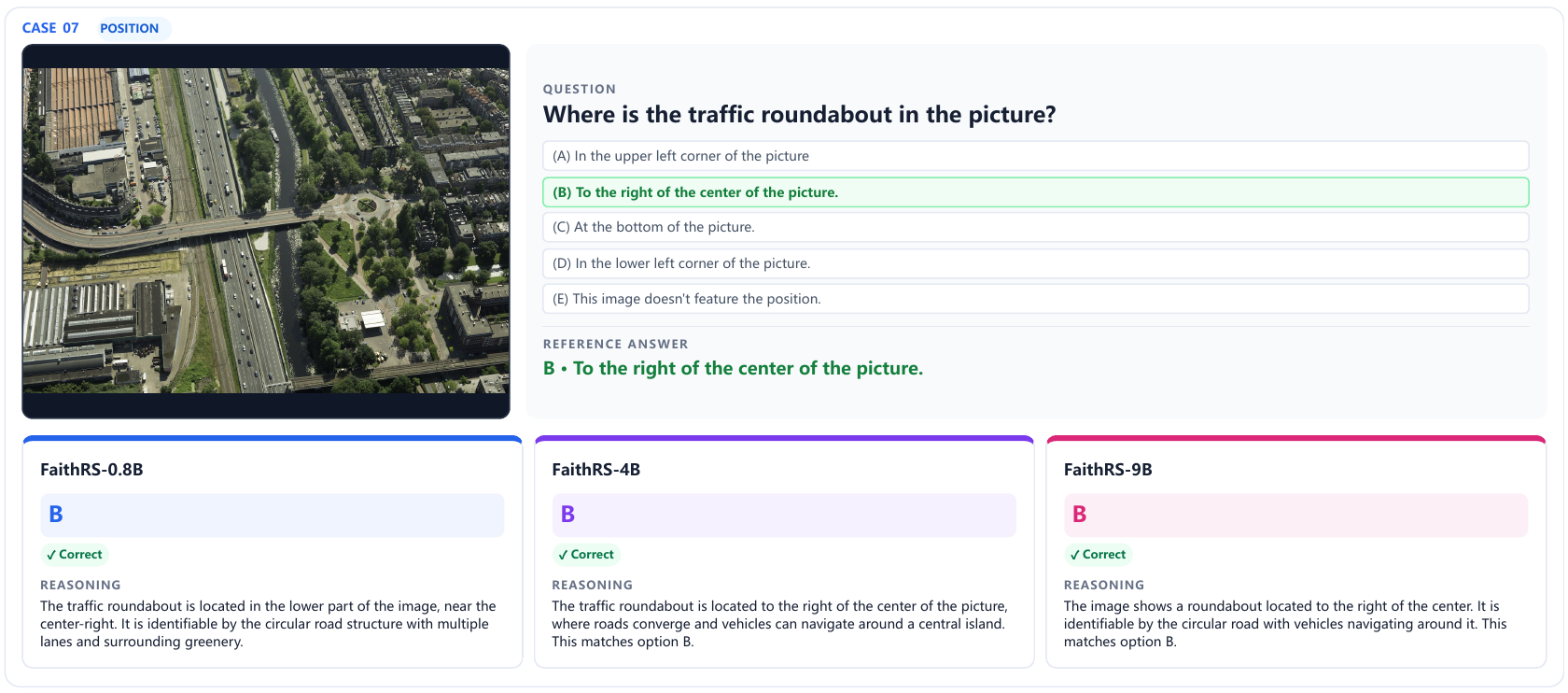}
\includegraphics[width=0.93\linewidth,height=0.27\textheight,keepaspectratio]{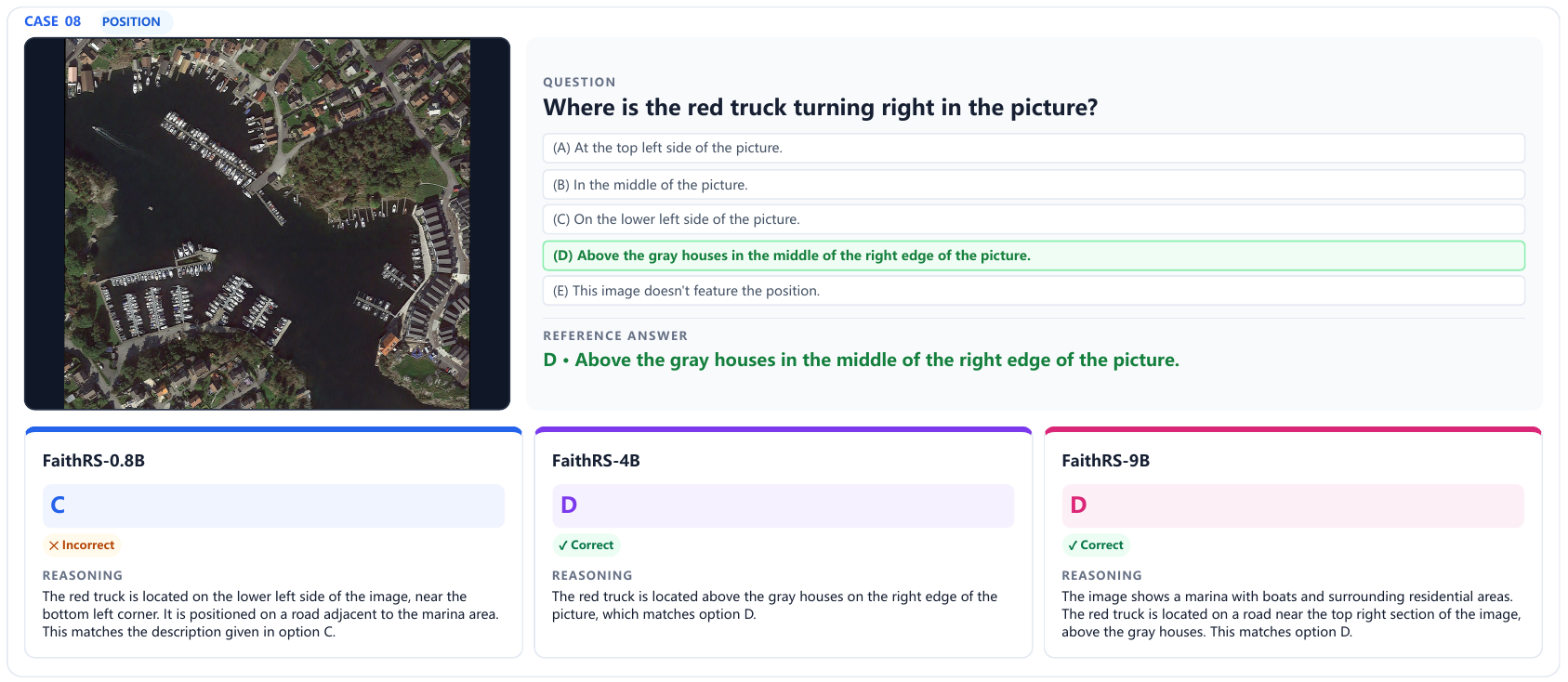}
\includegraphics[width=0.93\linewidth,height=0.27\textheight,keepaspectratio]{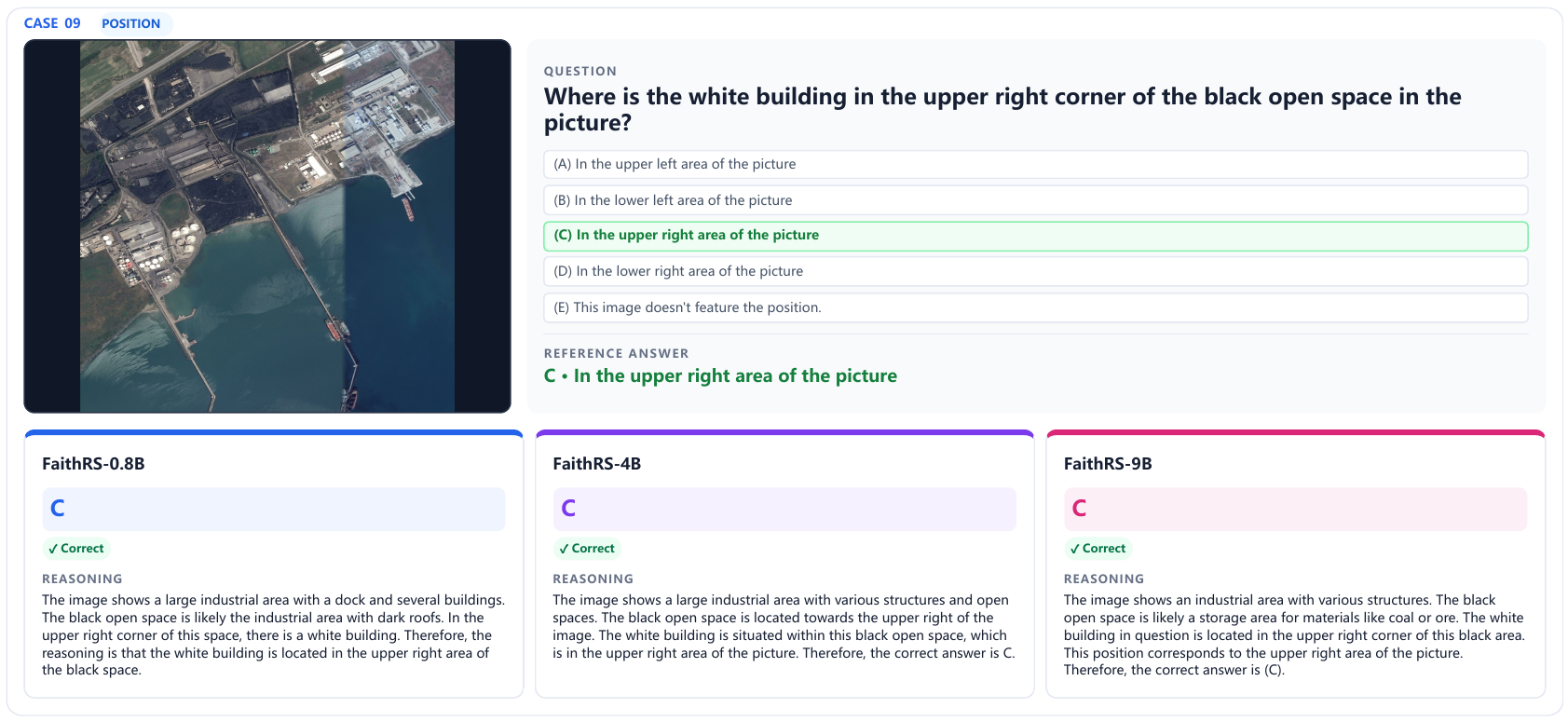}
\caption{Representative spatial-localization examples from MME-RealWorld-RS. Case 7 is answered correctly by all three FaithRS scales and provides a stable evidence-grounded localization example. Case 8 shows scale-dependent recovery from spatial misgrounding by FaithRS-0.8B. In Case 9, all three models select the correct location, although some explanations introduce semantic interpretations that are not directly verifiable from the image.}
\label{fig:qualitative_position}
\end{figure*}

\paragraph{Spatial localization and semantic faithfulness.} Figure~\ref{fig:qualitative_position} includes a stable localization success, a scale-dependent spatial correction, and a case in which correct answers coexist with unnecessary semantic interpretation. In Case 7, all three models correctly locate the roundabout to the right of the image center. Their explanations consistently refer to the visible circular road structure, providing a stable example of evidence-grounded spatial localization across model scales. In Case 8, FaithRS-0.8B places the red truck in the lower-left region near the marina, while FaithRS-4B and FaithRS-9B correctly locate it above the gray houses along the right edge. The smaller model produces a fluent and spatially specific description, but the described region is not grounded in the actual target location. Case 9 provides a complementary example in which all three models select the correct upper-right location of the white building. However, some explanations introduce additional interpretations of the surrounding black open space. In particular, FaithRS-9B describes the region as a storage area for coal or ore, although this functional interpretation is not required by the question and cannot be verified directly from the image. This case shows that correct spatial answers do not necessarily guarantee that every explanatory statement is visually supported.

Overall, the selected cases reveal three complementary properties of FaithRS. First, Cases 2, 6, and 7 demonstrate stable visual recognition, absence detection, and spatial localization across all three model scales, indicating that the learned corrective behavior does not prevent the models from retaining reliable task competence. Second, Cases 1, 4, 5, and 8 show that increasing model capacity can improve subtle attribute recognition, answer selection, small-target detection, and spatial grounding. Third, Cases 3 and 9 demonstrate that larger models remain vulnerable to shared perceptual errors and unsupported semantic elaboration. These qualitative results therefore complement the quantitative evaluation by showing both the reliable capabilities preserved by FaithRS and the reasoning failures that remain unresolved.

\clearpage
% Preserve the appendix bibliography records exactly as supplied in the original appendix.
\bibliographystyleapp{plainnat}
\bibliographyapp{appendix_references}

\end{document}